%% file: bare_jrnl.tex
\pdfoutput=1

\documentclass[journal]{IEEEtran}
\usepackage{amssymb}
\usepackage{algorithm}
\usepackage{algpseudocode}
\usepackage{color}
\usepackage{epsf}
\usepackage{psfrag}
\usepackage{epsfig}
\usepackage{graphicx}
\usepackage{amsfonts}
\usepackage{textcomp}
\usepackage{comment}
\usepackage{footnote}
\usepackage{tablefootnote}
\usepackage[flushleft]{threeparttable}
\usepackage{paralist}
\usepackage{mdwlist}
\usepackage{amssymb}
\usepackage{amsmath}
\usepackage{url}
\usepackage{verbatim}
\usepackage{alltt}
\usepackage{balance}
\usepackage{multirow}
\usepackage{epstopdf}
\usepackage{lipsum}
\usepackage{float}
\usepackage{booktabs}
\usepackage{slashbox}
\usepackage[table]{xcolor}
\usepackage{array}
\usepackage{boldline}
\usepackage{caption,setspace}
\usepackage{mathtools}

\usepackage{upgreek}
\usepackage{ stmaryrd }

\usepackage{calligra}
\usepackage[T1]{fontenc}
\usepackage{ upgreek }
\usepackage{ mathrsfs }

\usepackage[normalem]{ulem}
\usepackage{cancel}

\usepackage{cite}
\usepackage{amsmath}
\usepackage{graphicx}

\newcommand{\mhyphen}[1]{{\operatorname{#1}}}
\DeclareMathOperator{\E}{\mathbb{E}}

\ifCLASSINFOpdf
\else
\fi
\begin{document}
%
\title{Intra-Variable Handwriting Inspection \\Reinforced with Idiosyncrasy Analysis}
%
%
%

\author{Chandranath~Adak,~\IEEEmembership{Member,~IEEE,}
        Bidyut~B.~Chaudhuri,~\IEEEmembership{Life~Fellow,~IEEE,}
        Chin-Teng~Lin,~\IEEEmembership{Fellow,~IEEE,}
        and~Michael~Blumenstein,~\IEEEmembership{Senior~Member,~IEEE}
\thanks{C. Adak is with the Centre for Data Science, JIS Institute of Advanced Studies and Research, JIS University, India-700091, and also with School of Computer Science, FEIT, University of Technology Sydney, Australia-2007.  \newline 
(e-mail: chandra@jisiasr.org).
}
\thanks{B. B. Chaudhuri is with Dept. of CSE, Techno India University, India-700091, and also with CVPR Unit, Indian Statistical Institute, India-700108.}
\thanks{C.-T. Lin and M. Blumenstein are with the Centre for AI, School of Computer Science, FEIT, University of Technology Sydney, Australia-2007. }
\thanks{This paper is a preprint version of DOI: 10.1109/TIFS.2020.2991833}
}

\maketitle

\begin{abstract}
In this paper, we work on intra-variable handwriting, where the writing samples of an individual can vary significantly. Such within-writer variation throws a challenge for automatic writer inspection, where the state-of-the-art methods do not perform well. To deal with intra-variability, we analyze the idiosyncrasy in individual handwriting. We identify/verify the writer from highly idiosyncratic text-patches. Such patches are detected using a deep recurrent reinforcement learning-based architecture. An idiosyncratic score is assigned to every patch, which is predicted by employing deep regression analysis. For writer identification, we propose a deep neural architecture, which makes the final decision by the idiosyncratic score-induced weighted average of patch-based decisions. For writer verification, we propose two algorithms for patch-fed deep feature aggregation, which assist in authentication using a triplet network. The experiments were performed on two databases, where we obtained encouraging results.
\end{abstract}


%
\IEEEpeerreviewmaketitle

\input{1intro}

\input{1related}

\input{2idioAnal}

\input{3writerIdentification}

\input{4writerVerification}

\input{5experiment}

\input{6conclusion}

\bibliographystyle{IEEEtran} 
\bibliography{Bibliography} 




\begin{IEEEbiography}[{\includegraphics[width=1in,height=1.25in,clip,keepaspectratio]{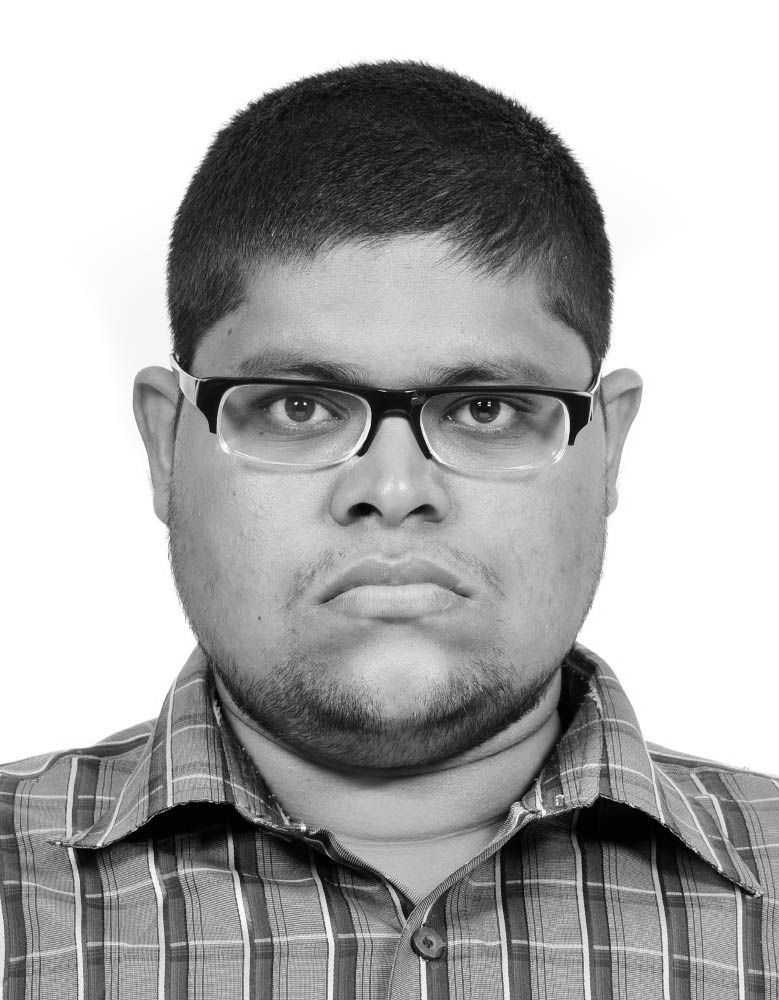}}]{Chandranath Adak} 
(S'13, M'20) received his PhD in analytics from University of Technology Sydney, Australia in 2019. Currently, he is an Assistant Professor at Centre for Data Science, JIS Institute of Advanced Studies and Research, India. His research interests include image processing, pattern recognition, document image analysis, and machine learning-related subjects.
\end{IEEEbiography}

\begin{IEEEbiography}[{\includegraphics[width=1in,height=1.25in,clip,keepaspectratio]{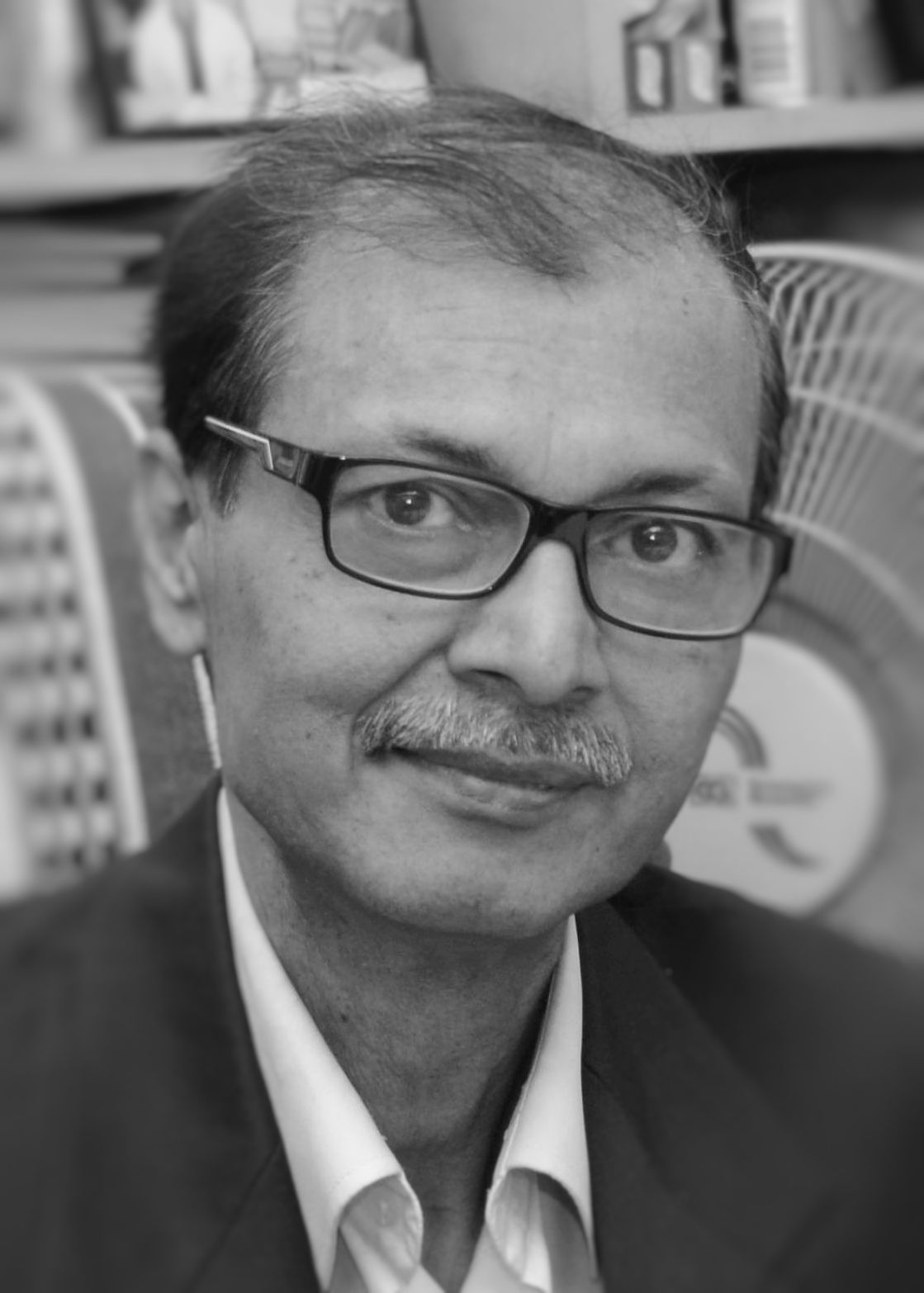}}]{Bidyut B. Chaudhuri}
(F'01, LF'16) received his PhD from Indian Institute of Technology, Kanpur in 1980. He joined the Indian Statistical Institute in 1978, and retired from the regular position in 2015. Currently, he is the Pro-Vice-Chancellor (Academic) of Techno India University, West Bengal, India. His research interests include pattern recognition, machine learning, digital document processing, natural language processing, speech processing, etc. 
\end{IEEEbiography}


\begin{IEEEbiography}[{\includegraphics[width=1in,height=1.25in,clip,keepaspectratio]{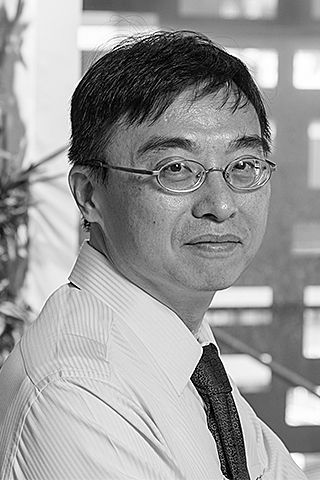}}]{Chin-Teng Lin} 
(F'05) received his PhD in electrical engineering from Purdue University, USA in 1992. 
He is currently a Distinguished Professor of CAI, FEIT, University of Technology Sydney, 
Chair Professor of Electrical and Computer Engg., NCTU, 
and Honorary Professor of University of Nottingham. 
His research interests include computational intelligence, fuzzy neural networks, brain-computer interface, robotics, etc.
\end{IEEEbiography}

\begin{IEEEbiography}[{\includegraphics[width=1in,height=1.25in,clip,keepaspectratio]{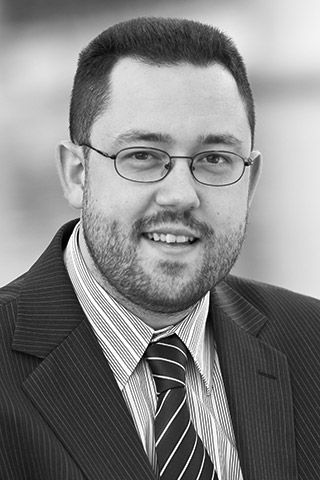}}]{Michael Blumenstein}
(SM'13) received his PhD in computational intelligence from Griffith University, Australia in 2001. He is currently a Professor and the Associate Dean (Research Strategy and Management) with the FEIT, University of Technology Sydney, Australia. His research interests include pattern recognition, artificial intelligence, video processing, document image processing, environmental science, neurobiology, coastal management, etc.
\end{IEEEbiography}


\newpage
\input{appendix}

\end{document}

%% file: 1intro.tex
\section{Introduction}
\label{intro}

\IEEEPARstart{H}{andwriting} is still considered as strong evidence in criminal courts of many countries due to its solid impact on behavioral biometrics \cite{forensic, 1}. Therefore, for the last four decades, research on handwriting inspection has been of great interest in forensics. Moreover, the computational approaches are embedded in handwriting forensics owing to booming automation since the late 20$^{th}$ century. Besides, the "9/11" and "2001 Anthrax" attacks have reignited the computational handwriting forensics research \cite{23}. 
	
From the forensic perspective, the handwritten specimen can be found mostly as an offline sample in the form of a threat letter, suicide note, forged manuscript, etc. \cite{forensic}. Therefore, in this paper, we focus on offline handwriting. The offline handwriting analysis is more challenging compared to online writing due to absence of stroke trajectory, writing pressure, velocity, etc.

In computational handwriting analysis, the focus during the last decade and the first half of the current decade were on handcrafted features. The deep neural net derived feature-based studies have thrived during the latter half of the current decade \cite{2}. Although the past researches on writer inspection have produced some encouraging results, the major works have been performed on inter-variable writing \cite{22}. The computational research on intra-variability of handwriting has been somewhat overlooked. However, handwriting intra-variability is observed rather frequently due to some mechanical, physical, and psychological factors of the writers \cite{10}. To the best of our knowledge, only one computational experiment has been performed on intra-variability due to Adak et al. \cite{10}. 
In that study, the authors experimentally showed that the general handcrafted and deep feature-based models did not work well on intra-variable writer inspection, i.e., training/testing on disparate writing styles. 
{
At this point, our paper comes into place to 
identify/verify 
the writer from intra-variable writing. 
Some important circumstances where our method is relevant are as follows.}

{
{\bf{\em(i)}} \emph{Absent data}: In forensics and biometrics \cite{forensic, factor}, for a writer examination system, a specific writing style/type of an individual may be absent during training. Now, the system may be required for testing on that particular type of writing.}

{
{\bf{\em(ii)}} \emph{Discovered manuscript}: In archival science and library science, the authorship is checked when an unpublished historical cultural manuscript is discovered. 
The manuscript may contain some unknown writing styles of a claimant \cite{factor}. 
A system may be essential to verify such a claimant.}

{
{\bf{\em(iii)}} \emph{Healthcare}: Some diseases, e.g., Parkinson's, Dyslexia, Alzheimer's, Dysgraphia, Tourette syndrome, etc., affect the handwriting of an individual \cite{parkinson, Alzheimer}. Therefore, the writing of a patient changes over multiple stages of disease progression. A system that can understand such intra-variability, may be needed to analyze such disease development. %
Our proposed system has the potential to address such real-world issues.}

\begin{figure}[!b]
\centering
\includegraphics[width=0.7\linewidth]{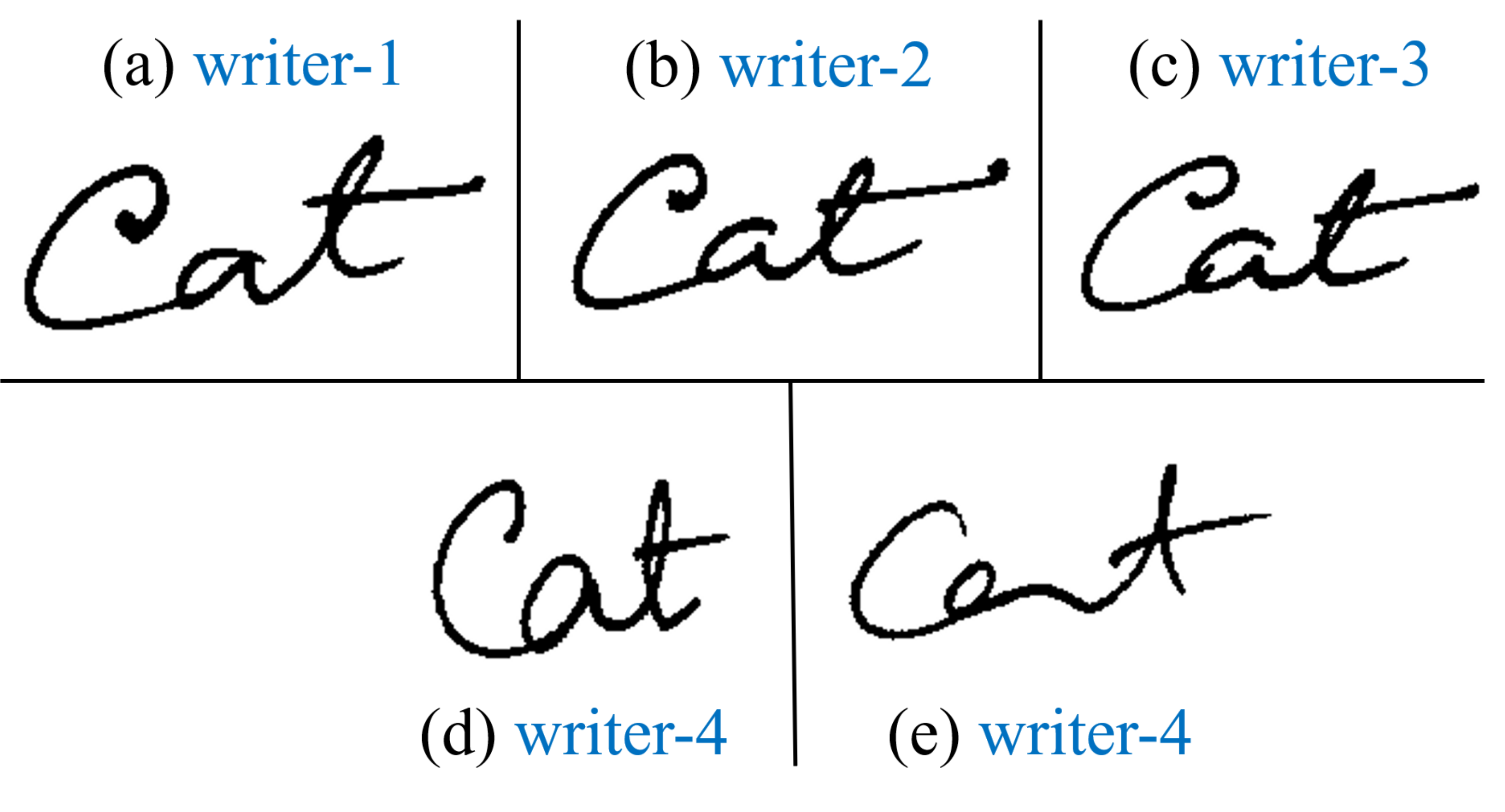}
\caption{
\footnotesize
(a), (b), (c): $3$ samples written by $3$ different writers: \emph{low inter-variability}; (d), (e): $2$ samples written by the same writer: \emph{high intra-variability}.
}
\label{fig:fig1}
\end{figure}

\begin{figure}
\centering
\includegraphics[width=0.75\linewidth]{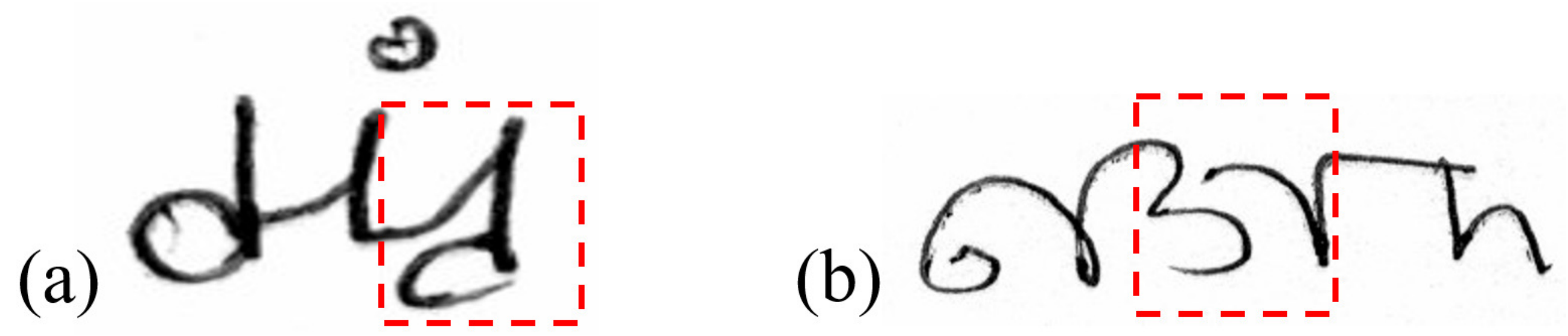}
\caption{
\footnotesize
Idiosyncratic writing samples in (a) \emph{English} and (b) \emph{Bengali} scripts, marked in {\color{red}{red}} dashed boxes: 
{
(a) eccentric cursive stroke to scribble character `d', 
(b) queer penning of the Bengali character containing intermittent stroke.}
}
\label{fig:fig2}
\end{figure}


In Fig. \ref{fig:fig1}, we present some examples on intra-variable and inter-variable handwriting. The samples of the upper row (Fig. \ref{fig:fig1}:(a)-(c)) appear to be structurally similar; however, these are written by three different writers. It depicts the \emph{low inter-variability} which is mostly performed with the intention of writing/signature-forgery \cite{24}. Here, writer-2 (Fig. \ref{fig:fig1}:(b)) and writer-3 (Fig. \ref{fig:fig1}:(c)) try to forge the inscription of writer-1 (Fig. \ref{fig:fig1}:(a)). In Fig. \ref{fig:fig1}:(d), (e), two writing samples appear to be dissimilar, but both are written by the same writer, i.e., writer-4. It portrays \emph{high intra-variability}. In this paper, we are concerned with such high intra-variability in contrast with the past works \cite{22, 23}.

{
For intra-variable handwriting inspection, the idiosyncrasy analysis \cite{3} of handwriting may be useful, 
since the forensic experts and paleographers follow quite a similar manual approach \cite{forensic}. 
}
Idiosyncrasy analysis of handwriting refers to examining the eccentricity in individual writing style. 
{
The originating Greek word of idiosyncrasy is "\emph{idiosunkrasia}", i.e. \emph{idios} (own, private)
+ \emph{sun} (with) + \emph{krasis} (mixture), which denotes the "distinctive or peculiar feature or characteristic"\footnote{https://en.oxforddictionaries.com , last retrieved on 30 March 2020.} 
of an individual. 
}
We observe that almost every writer scribbles some character-texts in a peculiar style, which may be useful to inspect the writer on intra-variable writing. In Fig. \ref{fig:fig2}:(a) and (b), we present two examples in English and Bengali scripts, where the writing idiosyncrasy is marked by red dashed boxes. Usually, to write the English character `d', at first the lower loop is scribbled, then the vertical straight line is drawn. However, in Fig. \ref{fig:fig2}:(a), to write `d', the vertical line is penned before the loop creation. Therefore, here, instead of the lower part (loop), the upper part (vertical line) of the `d' creates a continuity with the previous character, which represents the individual idiosyncrasy. In Fig. \ref{fig:fig2}:(b), to write the Bengali character `\includegraphics[height=7pt]{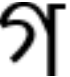}', an unnecessary ink-stroke gap makes the character penning highly {idiosyncratic (IdS)}.   
 
In \cite{3}, a preliminary work on idiosyncrasy analysis is performed, which did not deal with intra-variable writing; however, it has provided an insight that such analysis has a positive impact on Writer Identification {(WI)}. The authors modeled the idiosyncrasy analysis task into a classification problem to classify the text-patches into multiple classes defined by an {IdS} score. Their patch selection is mostly based on a sequential search with character-level information. In the current paper, we formulate the idiosyncrasy analysis task in a more sophisticated way, where we predict the score through deep regression \cite{25}, and select highly {IdS} patches using Reinforcement Learning ({RL}) \cite{5}. 

In this paper, we inspect the writer from the {IdS} patches, instead of using all the patches that was performed in \cite{10}. 
For writer inspection, a handwritten document is examined. 
In this paper, the examination involves the identification and verification of a writer.  
In the {WI} task, we find the correct writer-id of a questioned handwritten sample from multiple samples of different writers of a database. 
As a matter of fact, {WI} is a multi-class classification problem, where we need to find an unknown writer class among multiple writer classes. 
In the Writer Verification {(WV)} task, we authenticate an asked handwriting sample whether it has been written by a particular writer or not. Therefore, {WV} is a binary classification problem. 
For {WI} and  {WV}, we use some deep learning-based features. We perform the experiment on the database used in \cite{10}, which contains relatively high intra-variable Bengali offline handwriting. 
The outcome of our method is better than that presented in \cite{10} (refer to Section \ref{5Exp}).

The 
{contributions} 
of our current research are briefly mentioned as follows:

{{\bf{\em{(i)}}} The state-of-the-art methods including \cite{10} do not perform so well to inspect the writer on highly intra-variable handwriting. The method proposed in this paper performs better than the past techniques. Merging the idiosyncrasy analysis with intra-variable handwriting inspection is newly proposed here.

{{\bf{\em{(ii)}}}  We find highly {IdS} patches, and perform writer inspection on these patches only. To obtain an {IdS} score of a patch, we use a deep-feature induced regression analysis \cite{25}. For highly {IdS patch localization}, we employ {RL} \cite{5}. In {RL}, we propose a novel internal reward shaping function which is computed using the {IdS} score (refer to Section \ref{2Idiosyncrasy_Analysis}).   

{{\bf{\em{(iii)}}} For {WI}, combining the decisions obtained from individual patches is a new contribution, where the overall decision is made by the {IdS} score-fed weighted average of the individual patch-based decisions (refer to Section \ref{3WI}).

{{\bf{\em{(iv)}}} For {WV}, we propose two separate methods (MAF and XAF) for generating a combined page-level deep feature from multiple patch-level features (refer to Section \ref{4WV}).

The rest of the paper is organized as follows.
{Section \ref{related} explores the related work in the area.}
Section \ref{2Idiosyncrasy_Analysis} discusses our proposed method for idiosyncrasy analysis. 
Then Sections \ref{3WI} and \ref{4WV} describe our {WI} and {WV} methods. 
The following Section \ref{5Exp} deals with the experiments and results of our proposed method. 
Finally, Section \ref{6conclusion} concludes this paper.

%% file: 1related.tex
\section{{Related Work}}
\label{related}

{
To the best of our knowledge and online/offline searching capacity, there is no direct computational work on intra-variable handwriting inspection for WI/WV except \cite{10}, in the literature. 
The notable state-of-the-art methods \cite{22, 1989Plamondon} did not tackle the writer-inspection problem concerning within-writer variation, which is the main focus of this paper. 
Only Adak et al. \cite{10} reported this issue after performing an empirical study, where they showed that major handcrafted feature, and deep feature-based writer inspection models, did not perform well on intra-variable handwriting. 
The challenge here is the training and testing executed on disparate writing styles of an individual.}

{
However, in this section, we briefly mention some significant state-of-the-art writer inspection methods.
A comprehensive survey on writer inspection up to the year 1989 is reported in \cite{1989Plamondon}. The overview of the recent offline writer investigations can be found in \cite{22}.
The writer-inspection techniques can broadly be divided into two categories:
{\bf{\em(i)}} handcrafted feature-based models \cite{2002Srihari, 23, COLD, SIFT, 153Sabourin, LBP_variation, oBIF, KAS, VLAD}  
and 
{\bf{\em(ii)}} deep feature-based models \cite{CNN_Christlein, CNN_Tang_Wu, He_deep, CNN_Fiel, 10}.}

{
{\bf{\em(i)}} \emph{Handcrafted feature-based models}:
Among handcrafted feature-based methods, the macro-micro feature-based model of Srihari et al. \cite{2002Srihari} was well appreciated 
due to successful writer inspection on a U.S. population of 1500 from various demographics. Their macro features 
were mostly based on some statistical stroke information about entropy, slope, contour, text-line height, slant, etc. The micro features contained the chain code regarding gradient, structural and concavity-based stroke characteristics.  
Bulacu and Schomaker \cite{23} proposed some handcrafted textural and allographic features. 
Their contour-hinge textural feature is very popular in the community due to its good performance. Besides, the grapheme emission PDF (Probability Distribution Function) allographic feature works well for writer inspection.}  

{
He and Schomaker \cite{COLD} extracted curvature-free handcrafted features based on LBP (Local Binary Patterns)-runs over the image and line distribution (COLD: Cloud of Line Distribution) over dominant points of the contour.} 
{
Wu et al. \cite{SIFT} detected SIFT (Scale-Invariant Feature Transform) keypoints, and obtained their SDS (SIFT Descriptor Signature) and SOH (Scale and Orientation Histogram) to inspect the writer. 
%
Some other handcrafted features, e.g., 
LBP and its variants \cite{153Sabourin, LBP_variation}, 
LPQ (Local Phase Quantization) \cite{153Sabourin}, 
oBIF (oriented Basic Image Features) \cite{oBIF}, 
K-Adjacent Segments \cite{KAS}, 
Zernike moments encoded into VLAD (Vectors of Locally Aggregated Descriptors) \cite{VLAD}, etc. 
have been used in the literature.}

{
{\bf{\em(ii)}} \emph{Deep feature-based models}:
The deep features are mostly extracted by a CNN (Convolutional Neural Network).
Christlein et al. \cite{CNN_Christlein} produced the CNN-based local activation features and a GMM (Gaussian Mixture Model)-based supervector for writer inspection.} 
{
Tang and Wu \cite{CNN_Tang_Wu} extracted deep features using CNN and accomplished the writer-inspection task by employing a joint Bayesian technique. 
%
He et al. \cite{He_deep} obtained CNN-based features from a handwritten word image. Their main writer inspection task took the transfer learning benefit from another auxiliary task (e.g., WR: Word Recognition, WLE: Word Length Estimation, etc.), where adaptive convolutional layers were used during the transfer. 
%
Fiel and Sablatnig \cite{CNN_Fiel} used a CaffeNet for deep feature extraction and the nearest neighbor classifier for writer inspection. 
%
In \cite{10}, an empirical study was conducted where 
some major deep architectures, such as 
SqueezeNet, 
GoogLeNet, 
Xception Net,
VGG-16, 
ResNet-101, etc. were analyzed for deep feature extraction.}


%% file: 2idioAnal.tex
\section{Idiosyncrasy Analysis}
\label{2Idiosyncrasy_Analysis}

In this section, we perform idiosyncrasy analysis, where the objective is to find some highly idiosyncratic patches from a handwritten text sample, which can assist in writer inspection. 

\subsection{Idiosyncratic Opinion Score}
\label{idio_score}

Before finding the highly idiosyncratic ({IdS}) patches, we need to define an idiosyncrasy measure, based on which we can mark the respective patches as high or low. We adopt the idea of \cite{3} to define this measure, i.e., subjective opinion score \cite{19}. Now, we discuss the procedure to obtain the ground-truth score.

For ground-truthing, on a given text-patch ($p_t$), several handwriting experts provided their opinion scores 
($I_j^{(t)}$) within a continuous range of [$I_L,I_H$]; $I_L, I_H \in \mathbb{R}^+$. Here, we choose $I_L = 0$, ~$I_H = 10$. The arithmetic mean $(I_\mu^{(t)})$ of these scores is the  \emph{idiosyncratic opinion score} of a patch $p_t$, i.e., 
$I_\mu^{(t)}=\frac{1}{e} \sum_{j=1}^e I_j^{(t)}$; 
where $e > 1$ is the total number of experts who put score on a patch $p_t$. For our task, $e \geq 30$, i.e., at least $30$ experts put individual scores on a patch. Adak et al. \cite{3} partitioned the score range [$I_L,I_H$] into $n_I$ number of bins (classes) of equal width and modeled a classification task to find highly {IdS} patch classes. 
We work here differently by using regression analysis, where we predict the {IdS} score of a patch. In this paper, the score interval [$I_L,I_H$] is normalized into [$0, 1$] to produce normalized {IdS} score $i_t$ of patch $p_t$, i.e., 
$i_t=\frac{I_\mu^{(t)}-I_L}{I_H-I_L};~ 0 \leq i_t \leq 1$. A patch $p_t$ with $i_t = 1$ refers to the highest {IdS} patch, whereas $i_t = 0$ refers to the lowest one. As a matter of fact, on a page, multiple patches with the same {IdS} score may present. 
The score $i_t$ assists in automated detection of highly {IdS} patches, as well as in writer inspection, as described in the following subsections. 

\subsection{Detecting Idiosyncratic Patches}
A handwritten page is scanned to be used as an input image. Now, the task is to detect the {IdS} patches from the input image. We consider the problem as a decision making process where an agent interacts with a visual environment, viz., scanned handwritten page, to detect the target patches. 
At each timestep, the agent partially observes the input image and decides where to focus on the next timestep.

We cast this problem as a partially observable MDP (Markov Decision Process) since it allows the agent to make a decision through stochastic control in discrete time and the entire environment is unobserved by the agent in a particular step \cite{5}. We employ an {RL}-based agent here, which take action to learn a policy for maximizing the reward \cite{5}. The agent takes input of the state of current image status. MDP consists of a set of components, i.e., set of states of the current environment, set of actions to achieve the goal, and the reward to optimize decision strategy.
In this paper, the agent's task is to find a patch from a handwritten page that can be used for writer inspection. The agent will learn the policy through {RL} to find the highly {IdS} patches.

\begin{figure}[!b]
\centering
\includegraphics[width=\linewidth]{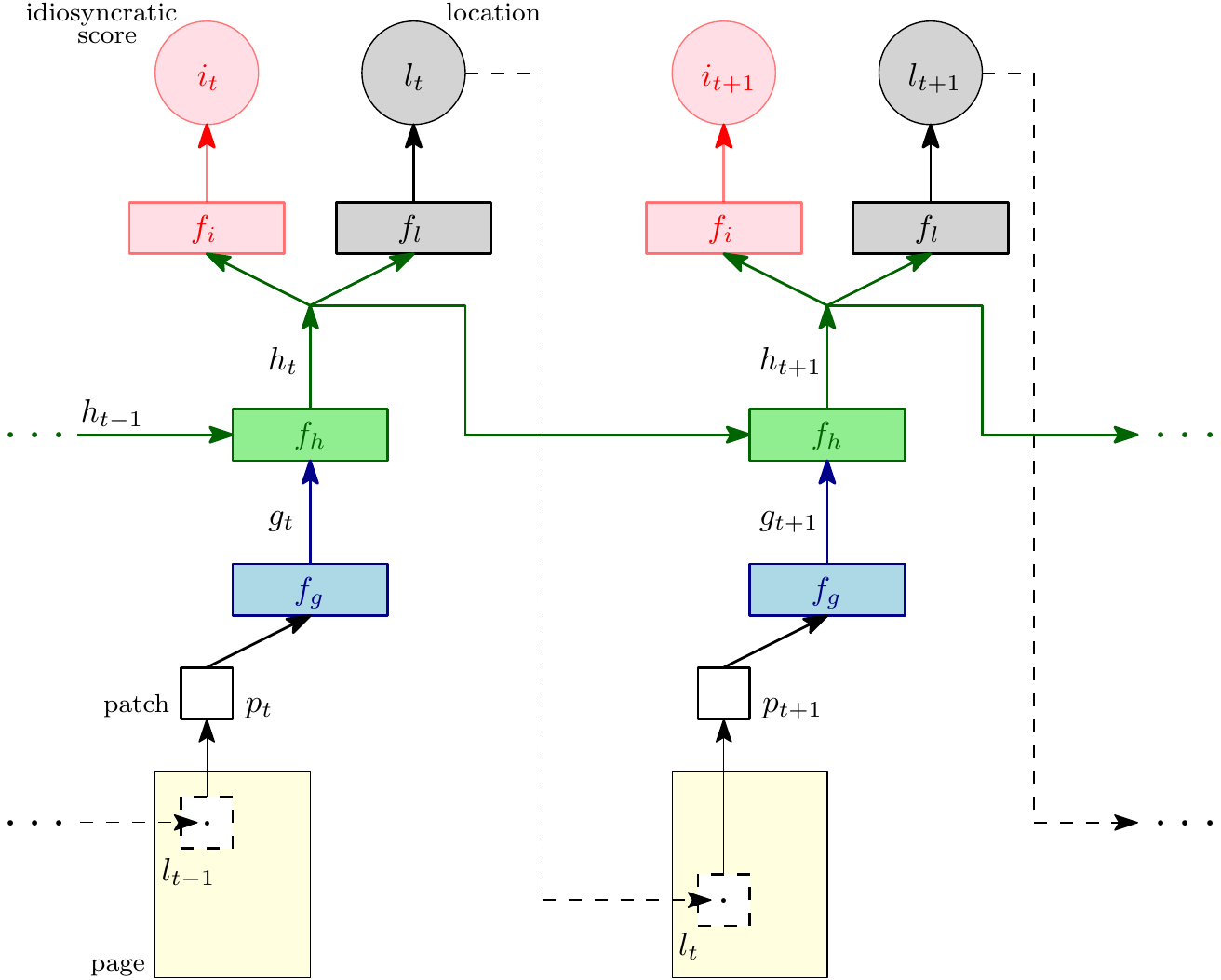}
\caption{Idiosyncratic patch detection.}
\label{fig:fig3}
\end{figure}
 

In a handwritten page, a text-patch ($p$) is a $w_p \times w_p$ square box centering at a location $l$ which is encoded with co-ordinate $(x,y)$. 
The co-ordinates of the whole page image is ranged between $(0,0)$ and $(1,1)$, where the top-left co-ordinate of the page is $(0,0)$ and bottom-right is $(1,1)$.

From a patch, we extract some deep neural network-based features.
{
For this, we use the ResNet-50 model \cite{4} which achieved human-like performance on ImageNet data \cite{13}, and also met our requirements. 
Besides, the skip connection concept of ResNet (Residual Network) makes the computation faster compared to some other deep architectures \cite{4, 13}, such as VGG-Net \cite{vgg}. 
The ResNet usually takes an input of size $224 \times 224$ \cite{4}. 
For our task also, we chose the $w_p$ equal to $224$.  
In Fig. \ref{fig:fig3}, $f_g$ is a ResNet-50 up to the "{avg pool}" layer \cite{4}, which produces} a $2048$-dimensional feature vector $g$. 
At timestep $t$, $g_t = f_g(p_t)$.


The feature $g_t$ is then fed into the core network $f_h$ which is basically an RNN (Recurrent Neural Network). We choose the RNN here, since it can memorize the prior patch information. Such memorization is crucial due to its impact on the current time step to find the next patch. 
The basic RNN unit, employed here is GRU (Gated Recurrent Unit) \cite{9} instead of LSTM (Long Short-Term Memory) due to its simplicity with similar performance gain for our task. 
%
GRU also attains lower computational cost since it has only $1$ internal state, and $2$ gates with fewer parameters, whereas LSTM has $2$ internal states and $3$ gates with more parameters. Our core network $f_h$ consists of $512$ GRU units. The current hidden state $h_t$ of RNN at timestep $t$ is a function of ResNet-produced feature $g_t$ and previous state $h_{t-1}$. It can be written using GRU gates as follows.
\begin{equation}
\begin{gathered}
h_t=f_h (h_{t-1},g_t )\\
or,~ h_t=\Gamma_u*c_t+(1-\Gamma_u )*h_{t-1}	
\end{gathered}
\end{equation}
where, $\Gamma_u=\sigma(linear(h_{t-1},g_t))$;  
$c_t = tanh(linear(\Gamma_r * h_{t-1},g_t))$; 
$\Gamma_r=\sigma(linear(h_{t-1},g_t))$.

$\Gamma_u$ and $\Gamma_r$ are two gates of GRU, i.e., \emph{update} and \emph{relevant} gates, respectively. Two types of activation functions, \emph{sigmoid} ($\sigma$) and $tanh$ are used in GRU. The $linear(\vartheta)$ represents a linear transformation of a vector $\vartheta$; i.e., $linear(\vartheta) = \varpi\vartheta + \flat,$ where, $\varpi$ is a weight matrix and $\flat$ is a bias vector.

Now, the $h_t$ is embedded to $f_i$ to predict an {IdS} score with respect to a textual patch. The $f_i$ contains a regression layer to generate a scalar-valued {IdS} score $i_t$, i.e., $i_t=f_i(h_t);~ 0 \leq i_t \leq 1$. 
The concept of linear regression on the top of a deep architecture is adopted here \cite{25}. The mean-squared-error is employed here as a loss function to train $f_i$, and the gradient is backpropagated through $f_h$ and $f_g$. The {IdS} score $i_t$ is used latter for reward shaping during {RL}.

The $h_t$ is also fed to $f_l$ for obtaining the next patch location $l$. In $f_l$, the policy for the location $l$ is decided by a $2$-component Gaussian with a fixed variance \cite{20}. The $f_l$ produces the mean of the location policy at time $t$, and is described as $f_l(h_t) = linear(h_t)$. Here $h_t$ denotes the state of the core network RNN. The $f_l$ is trained with {RL} to localize the next patch to focus.

In {RL}, an agent interacts with the state ($s$) of an environment and takes action ($a$) to obtain the reward ($r$) from the environment \cite{5}. In our task, the reward is generated internally at each time step $t$ instead of any environmental external reward. The state $s_t$ at $t$ takes patch input $p_t$ and summarized into internal state $h_t$ of RNN. The action $a_t$ at $t$ is actually the location-action $l_t$ selected stochastically from a distribution $\theta_l$-parameterized by $f_l(h_t)$ at $t$. In other words, the \emph{state} is the patches seen so far, and the \emph{action} is $(x,y)$ co-ordinate of the center of the next patch to be looked at.   

For \emph{reward} shaping, we propose an internal reward ($r_t$), generated from the {IdS} score $i_t$, as follows.
\begin{equation}
r_t  =
\begin{cases}
i_t  & \text{, if } i_t - i_{t-1} \geq T_{r1}~\text{and}~ i_t > T_{r2} \\
-(1-i_t)  & \text{, if } i_t - i_{t-1} < T_{r1}~\text{and}~ i_t > T_{r2}  \\
-i_t & \text{, otherwise}	
\end{cases}
\end{equation}
where, $T_{r1} > 0$ and $T_{r2} > 0$ are two thresholds.	
	
A positive internal reward ($r_t$) is provided here, if the {IdS} score ($i_t$) at current timestep $t$ has increased sufficiently from the score ($i_{t-1}$) of previous timestep ${t-1}$. For all other cases, we penalize the agent by providing a negative internal reward. For our task, $T_{r1} = 0.1$ and 
$T_{r2} = 0.5$ works well, that are set empirically.

An agent, in {RL}, entails to learn a stochastic policy $\pi_\theta(l_t | s_{1:t})$ with parameter $\theta$ at each timestep $t$, that maps the past trajectory of environmental interactions $s_{1:t} = p_1, l_1, \dots, p_{t-1}, l_{t-1}, p_t$ to an action distribution $l_t$. In our task, the policy $\pi_\theta$ is defined by the previously mentioned core network RNN, and $s_t$ is summarized by the state of $h_t$. For the parametrized policy $\pi_\theta$, the parameter $\theta$ is provided by the parameters $\theta_g$ and $\theta_h$ of the networks $f_g$ and $f_h$, respectively, i.e., 
$\theta = \{\theta_g, \theta_h\}$.
The agent learns parameter $\theta$ to find an optimal policy that maximizes the expected sum of discounted rewards ($r$). The cost function becomes as follows.
\begin{equation}
\small
J(\theta)=\E_{\rho(s_{1:T};~ \theta)}\left[\sum_{t=1}^T \gamma^{t-1} r_t \right]=\E_{\rho(s_{1:T};~ \theta)}\left[R\right]
\end{equation}
where, the transition probability $\rho$ from a state to another, depending on policy $\pi_\theta$ is specified as follows.
\begin{equation}
\small
\rho(s_{1:T};~ \theta) = \prod_{t=1}^T \rho(s_{t+1} | s_t, l_t) ~\pi_\theta(l_t | s_t)
\end{equation}
where, $T$ is the total count of time-step in an episode and $\gamma$ is a discounted factor.

Now, we find the optimal policy $\pi^*$ by optimizing the function parameter $\theta$. The optimal parameter $\theta^*$ is defined as $\theta^* = \underset{\theta}{\arg\max}~J(\theta)$. For finding the optimal policy, gradient ascent is used on policy parameters. Here, we borrow strategies from the {RL} literature \cite{6} as follows.
\begin{equation}
\small
\begin{aligned}
\nabla_\theta J(\theta) = \sum_{t=1}^T \E_{\rho(s_{1:T};~\theta)} \left[R~ \nabla_\theta \log \pi_\theta \left(l_t | s_t \right ) \right]\\ 
\approx
\frac{1}{N}\sum_{n=1}^N \sum_{t=1}^T R^{(n)} \nabla_\theta \log \pi_\theta \left(l_t^{(n)} | s_t^{(n)}\right)
\end{aligned}
\end{equation}
where, $s^{(n)}$'s are trajectories obtained by executing the agent on policy $\pi_\theta$ for $n=1, 2, \dots, N$ episodes. 
The gradient estimator does not depend on transition probability $\rho$. Moreover, the $\nabla_\theta \log \pi_\theta (l_t | s_t)$ part can be computed from the gradient of RNN with standard backpropagation \cite{7}.

The gradient estimator may suffer from high variance, therefore, variance reduction is necessary \cite{8}. 
The variance reduction with baseline ($b$) can be employed here to understand whether a reward is better than the expected one. Now, the gradient estimator takes the following form.
\begin{equation}
\small
\nabla_\theta J(\theta)
\approx
\frac{1}{N}\sum_{n=1}^N \sum_{t=1}^T \left(R_t^{(n)} - b_t \right) \nabla_\theta \log \pi_\theta \left(l_t^{(n)} | s_t^{(n)}\right)
\end{equation}
where, $R_t = Q^{\pi_\theta} (s_t, l_t) =  
\E\left[\sum_{t \geq 1} \gamma^{t-1} r_t | s_t, l_t, \pi_\theta \right]$
and 
$b_t = V^{\pi_\theta} (s_t) = \E \left[\sum_{t \geq 1} \gamma^{t-1} r_t | s_t, \pi_\theta \right]
$ are known as $Q$\emph{-value function} and \emph{value function}, respectively \cite{8}. The $Q$-value function follows the execution of action $l_t$, but the value function does not depend on $l_t$. We learn the baseline by reducing the squared error between $Q$-value function and value function.

In this paper, we adopt the idea of finding the next location through the recurrent neural network from \cite{20}. However, our architecture of Fig. \ref{fig:fig3} is quite new, where the $f_g$, $f_h$, $f_i$ nets are different from \cite{20}. The proposed internal reward-generating technique induced by {IdS} score is also a new contribution.

From the architecture of Fig. \ref{fig:fig3}, we obtain top-scoring $k$ number of idiosyncratic ({IdS}) patches. Therefore, the number of timesteps ($T$) in an episode equals to $k$. We also empirically fix the number of episodes ($N$) as $1000$.

%% file: 3writerIdentification.tex
\section{Writer Identification}
\label{3WI}

\begin{figure*}
\centering
\includegraphics[width=0.8\linewidth]{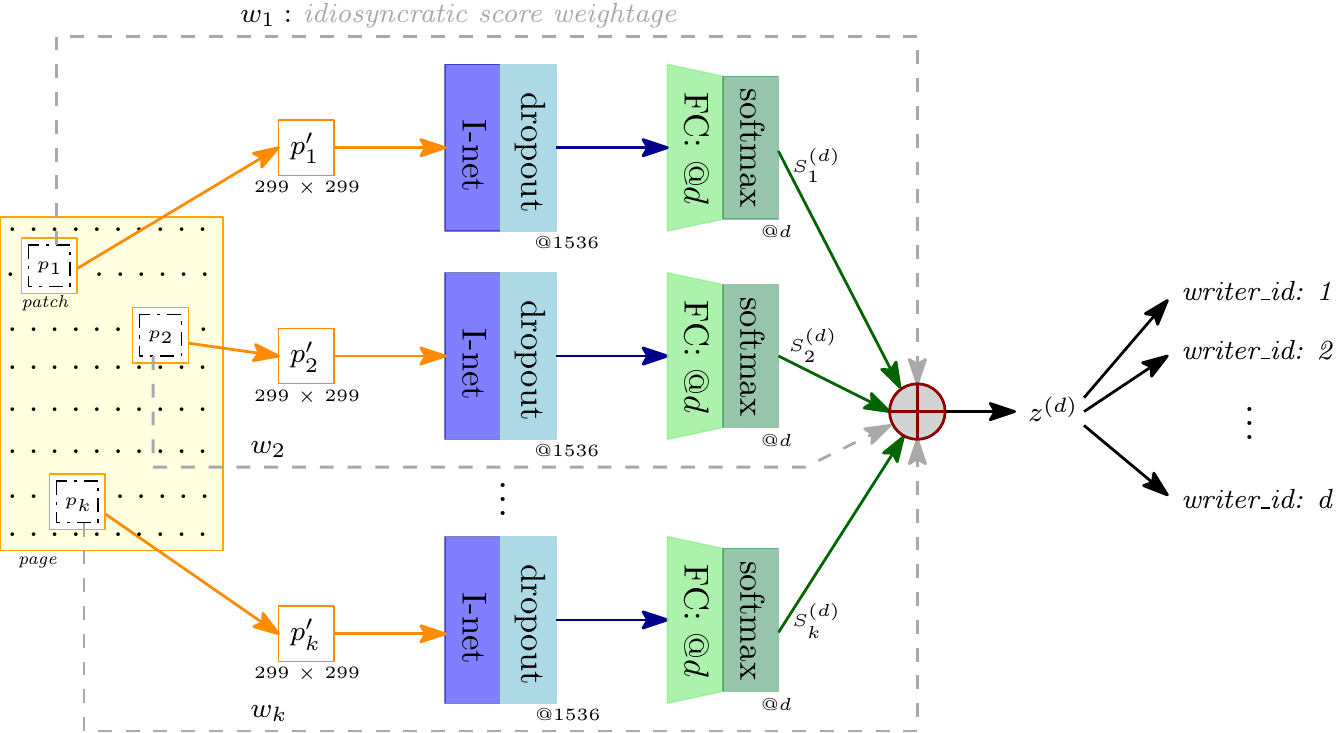}
\caption{Writer identification architecture.}
\label{fig:fig4}
\end{figure*}

From a handwritten page sample, we obtained $k$ number of highly {IdS} patches ($p_j; ~\text{for}~ j=1, 2, \dots , k$) that are used for writer identification ({WI}). 
Adak et al. \cite{10} 
showed promising outcomes from auto-derived features compared to handcrafted features while dealing with intra-variable writing. Therefore, in this paper, we focus on obtaining auto-derived deep features. Moreover, 
in \cite{10}, the authors
performed an empirical study with several state-of-the-art deep neural nets and obtained the best performance using Xception net \cite{11}. The contemporary Inception-ResNet-v2 architecture works better than Xception net and Inception-v4 \cite{11, 12} on ILSVRC database \cite{13}. 
Therefore, we adopt the Inception-ResNet-v2 \cite{12} for deep feature extraction only, from a handwritten text patch. 
The rest of the writer-inspection architecture is our proposal.

The size of an obtained patch $p_j$ is $224 \times 224$. We also attained the center location $l_j$ (south-east co-ordinate among four central locations) corresponding to each $p_j$. 
{From $p_j$}, we obtain $p_j'$ of size $299 \times 299$, by a padding of the proper width. The west and north sided padding widths are of size $\lfloor(299-224)/2\rfloor = 37$ each, whereas the east and south sided are of $\lceil(299-224)/2\rceil= 38$. 
The padded region is filled with the original intensity values of the {input image}. 
{We generate $p_j'$ to employ} some earlier layers of Inception-ResNet-v2 architecture {that usually} takes input of size $299 \times 299$.
We use up to the "average pooling" \cite{12} layer of Inception-ResNet-v2 as a feature extractor and refer to it as "I-net" in the rest of this paper. Therefore, I-net produces a $1536$-dimensional feature vector \cite{12}.

Now, we discuss the architecture for {WI} as shown in Fig. \ref{fig:fig4}. The last layer of the I-net is the "average pooling" \cite{12} layer of Inception-ResNet-v2. After this layer, we use a dropout \cite{14} of $20\%$ neurons to reduce over-fitting. Next, we add a fully connected (FC) layer to obtain a feature vector of size $d$ from each patch. Then this feature vector is transferred through a \emph{softmax} activation function. Each patch $p_j'$ produces a softmax probability distribution $s_j^{(d)};~ \forall j = 1, 2, \dots k$, over class labels.
$ \sum_d s_j^{(d)} =1 $.
Here, $k$ is the number of patches obtained from a page, and $d$ is the total number of classes, i.e., the total count of writers in a database. Now, all $s_j^{(d)}$'s obtained from $p_j'$'s are combined to obtain the $writer\_id$ of a page, as follows.
\begin{equation}
writer\_id = \underset{{d}}{\arg\max}~ z^{(d)} 
\end{equation}
where, $ z^{(d)} = \frac{\sum_{j=1}^k w_j s_j^{(d)}}{\sum_{j=1}^k w_j}$, and $  \sum_d z^{(d)}  = 1 $.
Here, $z^{(d)}$ is the weighted arithmetic mean of $s_j^{(d)}$'s. 
A weight $w_j$ is associated with $s_j^{(d)}$, which is determined from the {IdS} score $i_j$ of a patch $p_j$ as follows.
\begin{equation}
w_j=
\begin{cases}
\alpha_w i_j  & \text{, if } i_j > 0.1 \\
1 & \text{, otherwise}	
\end{cases}
\end{equation}
where, $\alpha_w$ is a scalar multiplier, empirically set to $10$.  


We use cross-entropy \cite{26} as the loss function because of its good performance in multi-class classification. 
{
The regularization \cite{26} is also used here to reduce the overfitting problem. 
The SGD (Stochastic Gradient Descent) with momentum is employed to optimize the cost function \cite{26}.}

%% file: 4writerVerification.tex
\section{Writer Verification}
\label{4WV}

In case of writer verification ({WV}), we authenticate an unknown handwritten sample based on the samples of a known writer-database. Therefore, here the task is to take input of two writing samples and produce the output either "\emph{same}" if they are written by the same writer, or "\emph{different}" otherwise \cite{23}. To measure the similarity between handwritten pages, we extract the page-level feature vectors corresponding to these pages, and compare between the feature vectors.  

\begin{figure*}
\centering
\includegraphics[width=\linewidth]{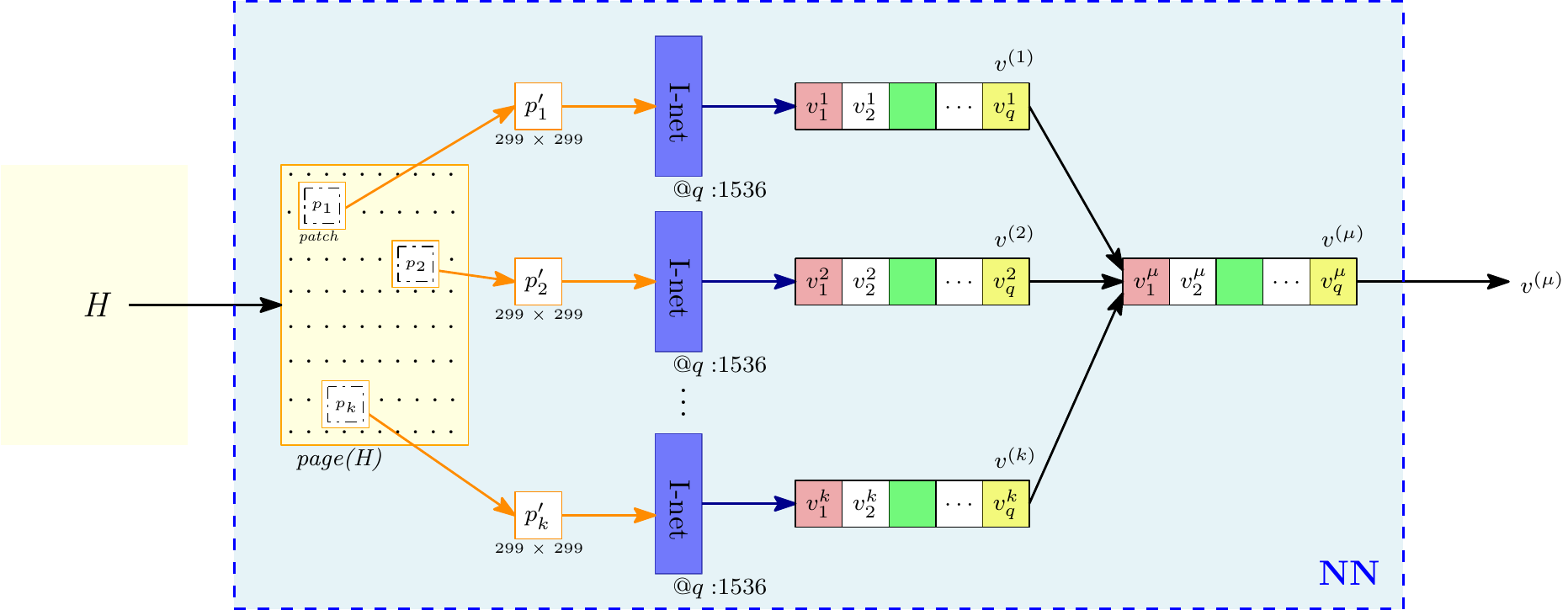}
\caption{Feature extraction for writer verification.}
\label{fig:fig5}
\end{figure*}

{The auto-derived deep features are also used here} 
since such features outperformed the handcrafted features \cite{10}. Similar to the case of {WI}, at first, we extract $q$ ($=1536$)-dimensional deep feature vectors 
($v^{(j)}: \{v_1^j,v_2^j, \dots,v_q^j\}; ~\forall j = 1, 2, \dots k $) 
from each of the top $k$ {IdS} patches ($p_j'$) using I-net (refer to Fig. \ref{fig:fig5}). Now, all the feature vectors $v^{(j)}$'s obtained from all the patches $p_j'$'s are aggregated to obtain a single $q$-dimensional feature vector ($v^{(\mu)}: \{v_1^\mu,v_2^\mu, \dots, v_q^\mu \}$) corresponding to a handwritten page sample ($H$). 
{
The $v^{(\mu)}$ is calculated as follows. 
$v^{(\mu)}: 
\{
v_1^\mu =  \frac{1}{k}\sum_{j=1}^k v_1^j, ~
v_2^\mu =  \frac{1}{k}\sum_{j=1}^k v_2^j, ~
\dots, ~
v_q^\mu =  \frac{1}{k}\sum_{j=1}^k v_q^j
\}$. 
%
%
%
%
We present this method of generating a \emph{Mean Aggregated Feature} (MAF) from multiple text-patches of a page in Algorithm 1: MAF 
(refer to Appendix A of the supplementary material).
Our MAF algorithm is different from the page-level feature generation using \emph{Strategy-Mean} of \cite{10}.
}



{
We propose another \emph{maX Aggregated Feature} (XAF) generation from multiple patches of a page in Algorithm 2: XAF (refer to Appendix A of the supplementary material).} 
The aggregated feature $v^{(\mu)}$ is calculated as follows. 
$ v^{(\mu)}: \{v_1^\mu=\max(v_1^1,v_1^2, \dots, v_1^k), 
~v_2^\mu=\max(v_2^1,v_2^2, \dots, v_2^k), 
~\dots, 
~v_q^\mu = \max(v_q^1, v_q^2, \dots, v_q^k) \}$. 
Our XAF algorithm is also different from \emph{Strategy-Major} of \cite{10} for page-level feature generation.

\begin{figure}
\centering
\includegraphics[width=\linewidth]{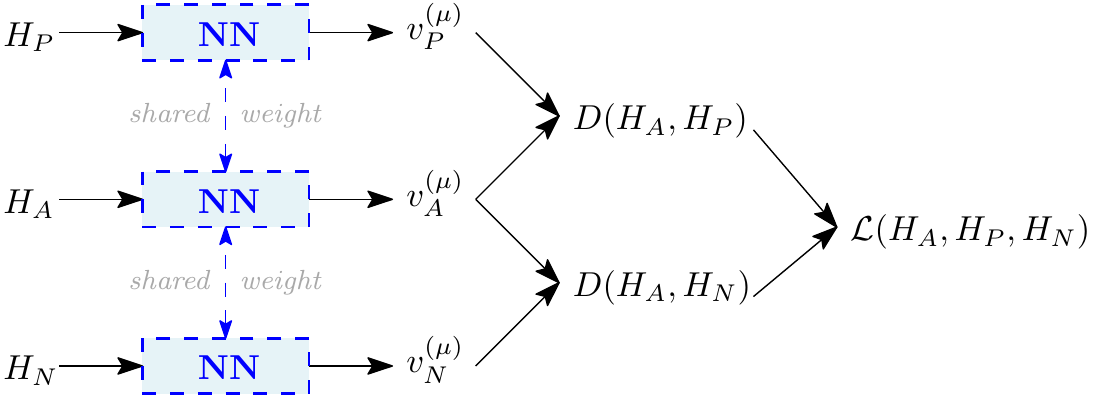}
\caption{Triplet Network (refer to Fig. \ref{fig:fig5} for NN).}
\label{fig:fig6}
\end{figure}

Thus, we have obtained a feature-vector $v^{(\mu)}$ from a handwritten page ($H$). For {WV}, we need to examine the writing style similarity/dissimilarity between pages. In other words, we measure the similarities among feature-vectors $v^{(\mu)}$'s representing pages $H$'s. 
For similarity metric learning, we adopt the idea of triplet network \cite{16}, 
{
which is quite a popular technique in the literature \cite{15, 16, 17}.}  
In the triplet net (refer to Fig. \ref{fig:fig6}), three identical neural nets (NN's) produce three feature vectors $v_A^{(\mu)}, v_P^{(\mu)}, v_N^{(\mu)}$ in parallel from three handwritten pages $H_A, H_P, H_N$, respectively, i.e., $v_A^{(\mu)}=NN(H_A), ~v_P^{(\mu)}=NN(H_P), ~v_N^{(\mu)}=NN(H_N)$. The three NN's share weights among them.

In the triplet network, we compare a positive sample ($H_P$) and a negative sample ($H_N$) with reference to an anchor/baseline sample ($H_A$), simultaneously. $H_A$ and $H_P$ handwritten samples are written by the same writer, whereas $H_A$ and $H_N$ samples are written by two different writers. We use Euclidean distance ($D$) between $v_A^{(\mu)}$ and $v_P^{(\mu)}$ as a distance metric to compare $H_A$ and $H_P$, i.e., 
$D(H_A, H_P) = {\parallel v_A^{(\mu)} - v_P^{(\mu)} \parallel}_2$. 
Similarly, we compare $H_A$ and $H_N$ with 
$D(H_A, H_N) = {\parallel v_A^{(\mu)} - v_N^{(\mu)} \parallel}_2$.
Finally, the wings of the triplet network are joined using a loss function, called triplet loss (${\cal{L}}$) \cite{15} to train with the similarity/dissimilarity metric. 
This ${\cal{L}}$ is defined as follows.
\begin{equation}
\footnotesize
    {\cal{L}}(H_A, H_P, H_N) = \max \left( D(H_A,H_P) - D(H_A,H_N) + \alpha_m , ~0 \right) 
\end{equation}
where, $\alpha_m$ is a margin parameter. Empirically, $\alpha_m$ is set as $0.2$, when checked in the interval $[0.1, 0.9]$ with a step of $0.1$.
This triplet loss ensures the positive sample to be closer to the anchor than that of the negative one, by at least a margin $\alpha_m$.

The overall cost function (${\cal{J}}$) of the triplet network is the sum {of individual losses on different triplets, over the training set cardinality $M$,} which is given as follows.
\begin{equation}
\small
{\cal{J}} = \sum_{m=1}^M {\cal{L}} \left( H_A^{(m)}, H_P^{(m)}, H_N^{(m)} \right)
\end{equation}
Here, with reference to an 
{
anchor sample ($H_A$), we choose the hardest positive sample ($H_P$) and the hardest negative sample ($H_N$) within a mini-batch to form a triplet \cite{17}. 
This hard-triplet is hard to train, which} increases the computational efficiency of the learning algorithm. 
The SGD 
 with momentum is employed here for minimizing ${\cal{J}}$.

In \cite{10}, a Siamese net with the contrastive loss \cite{18} is used for {WV}. In this paper, we utilize a triplet loss-based network, since it works better than the contrastive loss-based Siamese net \cite{17, 18, 15}.

All handwritten page pairs ($H_i, H_j$) scribbled by the \emph{same} writer are represented by ${\cal{P}}_{same}$, and all writing sample pairs of different writers are denoted as ${\cal{P}}_{diff}$. For system performance evaluation, we define a set of true positives ($TP$) at a threshold $t_d$, when all writing sample pairs are correctly classified as "\emph{same}", i.e., 
\begin{equation}
\small
TP(t_d) = \{(H_i,H_j) \in {\cal{P}}_{same}, ~\text{with}~ D(H_i, H_j) \leq t_d\}    
\end{equation}

Similarly, a set of true negatives ($TN$) at $t_d$ is defined where all pairs are correctly classified as "\emph{different}", i.e.,
\begin{equation}
\small
TN(t_d) = \{(H_i,H_j) \in {\cal{P}}_{diff}, ~\text{with}~ D(H_i,H_j) > t_d\}
\end{equation}

The true positive rate ($TPR$) and true negative rate ($TNR$) for a given writing distance $t_d$ are defined below.
\begin{equation}
\small
TPR(t_d ) = \frac{| TP(t_d) |}{| {\cal{P}}_{same} |} ; ~~ 
TNR(t_d ) = \frac{| TN(t_d) |}{| {\cal{P}}_{diff} |}   
\end{equation}

The overall \emph{accuracy} (balanced) \cite{21} for {WV} is calculated as follows.
\begin{equation}
\small
Accuracy = \max_{t_d \in D} \frac{TPR(t_d) + TNR(t_d)}{2} 
\end{equation}
where, $t_d$ varies with a step of $0.1$ in the range of $D$.


%% file: 5experiment.tex
\section{Experiments and Discussions}
\label{5Exp}

In this section, we discuss the experiments performed to evaluate our proposed system. We analyzed the system performance based on idiosyncratic patch detection, writer identification ({WI}), and writer verification ({WV}). We also compared the proposed approach with some past methods. Before proceeding to the experimental analysis, we first present the database employed for our experiments.

\subsection{Database Employed} \label{DB_emp}

For experimental analysis, we required a database (DB) containing intra-variable handwritten pages of a writer.
{
As far as we know, in the literature, except \cite{10}, 
no database contains 
such intra-variable handwritten samples. 
Therefore, we only used the} two databases of \cite{10}, namely \emph{controlled} ($D_c$) and \emph{uncontrolled} ($D_{uc}$) datasets which are briefly discussed below.

\subsubsection{Controlled ($D_c$)}
This database comprises a total of $600$ Bengali handwritten pages written by $100$ writers, i.e, $6$ pages per writer. This database contains $3$ sets ($S_f, S_m, S_s$) of intra-variable writing 
{based on various speeds of writing ($f$ast, $m$edium, $s$low).}  
Each of these $3$ sets has $2$ handwritten pages per writer. 
For example, a writer's handwritten sample of $S_f$ set varies extensively with his/her writing sample of $S_s$.  

{
$D_c$ was generated in a controlled way, where the writers were aware of the experimentation and were asked to write at different speeds, e.g., normal/medium, faster than normal, and slower than normal. 
To validate the writing speed ($\upnu_w$), the writing time ($\uptau$) of a page was noted. The total stroke-length ($\ell$) of a page was also computed \cite{10}. Now, $\upnu_w = \ell / \uptau$. 
The decision to place a handwritten page into one of the subsets $S_f, S_m, S_s$ was taken based on $\upnu_w$ with respect to two data-driven thresholds.
More details on $D_c$ can be found in \cite{10}.}

\subsubsection{Uncontrolled ($D_{uc}$)}
Similar to $D_c$, this database contains $600$ Bengali handwritten pages of $100$ individuals, where each writer wrote $6$ pages. 
{However, the $D_c$ and $D_{uc}$ datasets are writer-disjoint.} 
Here too, the database is divided into $3$ sets ($S_f’, S_m’, S_s’$) of intra-variable writing, where each set comprises $2$ pages per writer. 

{
The $D_{uc}$ was created in an uncontrolled manner, where the writers were not informed of the experimentation before their scribbling.
For this, a real school examination was conducted. 
In the answer sheets, intra-variation was observed due to the writing speed, and some behavioral/psychological aspects \cite{Behavior} such as anxiety, stress to recall answers, nervousness to complete, the panic of low marks, etc. 
The handwritten pages were grouped into $S_f’, S_m’, S_s’$ sets using a deep feature-based clustering technique followed by validation of handwriting experts. An elaborate discussion on $D_{uc}$ is reported in \cite{10}.}


The data of $D_c$ and $D_{uc}$ were augmented 
{to reduce the overfitting problem.  
In} data augmentation, the \emph{offline DropStroke} \cite{10} technique was used, where some writing-strokes were dropped randomly without creating any extra stroke components. A graph-based strategy was used here to obtain the strokes.
{
A method as given in \cite{SignAugment, DiazSignAug2} that generates duplicate offline synthetic signatures using some cognitive-inspired algorithm, may be an alternative data augmentation technique, but a costly process for producing full-page writing instead of a signature.} 

{
For data augmentation \cite{10}, a full handwritten page was roughly split horizontally into two halves \cite{SIFT}. 
Then on each half, the \emph{offline DropStroke} was used to generate 10 augmented samples. 
Therefore, a full page generated $22~(= 2 \times (1+10))$ samples. 
Now, each of the sets $S_f$, $S_m$, $S_s$ of $D_c$ and $S_f’$, $S_m’$, $S_s’$ of $D_{uc}$ contained 2 full pages per writer, therefore, each of them generated $44~(= 2 \times 22)$ samples from each of the $100$ individuals. 
Each of these samples was input to our system.}

	
In this paper, the experimental setup was similar to that in \cite{10}. Both the databases $D_c$ and $D_{uc}$ were divided into training, validation, and test set in the ratio of $2:1:1$. 
For example, $S_f$ set was divided into $S_{f1}$ (training), $S_{f2}$ (validation), and $S_{f3}$ (test) subsets, which contained $22$, $11$, and $11$ handwriting samples from each of the $100$ writers.
{
Actually, $S_{f1}$ contained samples obtained from the first full page of $S_{f}$, whereas, 
$S_{f2}$ and $S_{f3}$ comprised samples acquired from top and bottom halves of the second page of $S_{f}$, respectively.
}
As a matter of fact, 
$S_f = S_{f1} \cup S_{f2} \cup S_{f3}$. 
Similarly, 
$S_m = S_{m1} \cup S_{m2} \cup S_{m3}$, 
$S_s = S_{s1} \cup S_{s2} \cup S_{s3}$, 
$S'_f = S'_{f1} \cup S'_{f2} \cup S'_{f3}$, 
$S'_m = S'_{m1} \cup S'_{m2} \cup S'_{m3}$, and 
$S'_s = S'_{s1} \cup S'_{s2} \cup S'_{s3}$.

\subsection{{Performance of} Idiosyncratic Patch Detection}
\label{perf_idio}

In this subsection, we analyze the performance of our system for detecting idiosyncratic ({IdS}) patches.

{
As mentioned in Section \ref{idio_score}, the handwriting experts were given multiple patches for ground-truthing. 
The patches were obtained by sliding a $224 \times 224$ sized window horizontally through the text-lines with a stride of $96$, fixed empirically. 
Furthermore, the experts were shown a full page of handwriting on which they pointed to some patches randomly and put their opinion scores. 
The experts provided their scores independently without knowing the scores provided by others.}  

The proposed system predicted the {IdS} opinion score of a detected patch through deep regression analysis. Therefore, we used a standard performance measure for prediction, i.e., MAE (Mean Absolute Error). MAE is the arithmetic mean of the absolute differences between actual and predicted {IdS} opinion scores of the patches. 
%
%
{
The actual IdS score was obtained by averaging the scores of those patches with whom the IoU (Intersection over Union) \cite{iou} measures of the predicted patch were greater than 0.5.} 
The employed training, validation, and test sets are discussed in the previous subsection. Some hyper-parameters were also tuned here.

In TABLE \ref{tab:table1}, we present the MAE results when $k$ number of patches were chosen from each text sample. The results obtained from both $D_c$ and $D_{uc}$ databases are shown here. For $k = 125$, we obtained the lowest MAE of 1.03\% and 1.67\% for $D_c$ and $D_{uc}$, respectively. 

MAE focused on measuring the correctness of predicting the {IdS} score and did not guarantee to infer a \emph{highly} {IdS} opinion score. However, in this research, we were interested in highly {IdS} patches for writer inspection. Therefore, it was necessary to propose some measure which focused on analyzing highly {IdS} opinion score.

At this point, we proposed a performance measure, which was the arithmetic mean of normalized {IdS} opinion scores of $k$ number of patches, which were obtained from each text sample. This {Mean Idiosyncratic Score} (MIS) inferred the detection of highly {IdS} patches, when the MIS was high. 
As shown in TABLE \ref{tab:table1}, we obtained the highest MIS of 0.879 and 0.868 for $D_c$ and $D_{uc}$, respectively, when $k = 100$. However, the MIS did not guarantee the correct prediction of an {IdS} score. Therefore, to analyze the correctly-predicted highly {IdS} patch, we intended to observe both the MAE and MIS. 
{From TABLE \ref{tab:table1}, we can see that} 
$k=125$ produced the lowest MAE, whereas $k=100$ produced the highest MIS. Therefore, we checked the writer-inspection performance by varying the value of $k$ in the following subsections. 


\begin{table}
\scriptsize 
\caption{\small Performance of idiosyncratic patch detection}
\centering
\begin{tabular}{c|c|c|c|c}
\hline
\multirow{2}{*}{\textbf{\# patches}} &	\multicolumn{2}{c|}{\textbf{Mean Absolute}}  &	\multicolumn{2}{c}{\textbf{Mean Idiosyncratic}} \\ \multirow{2}{*}{ ($k$)}        &	\multicolumn{2}{c|}{\textbf{Error (MAE) \%}}  &	\multicolumn{2}{c}{\textbf{Score (MIS)}} \\  \cline{2-5}
 &	$D_c$ &	$D_{uc}$ &	~$D_c$~ &	$D_{uc}$ \\ \hline \hline
50	& 2.08	& 2.96	& 0.618	& 0.610 \\ \hline
75	& 1.77	& 1.97	& 0.774	& 0.761 \\ \hline
100	& 1.30	& 2.08	& \textbf{0.879}	& \textbf{0.868} \\ \hline
125	& \textbf{1.03}	& \textbf{1.67}	& 0.832	& 0.826 \\ \hline
150	& 1.75	& 1.86	& 0.783	& 0.779 \\ \hline
175	& 2.14	& 2.02	& 0.658	& 0.651 \\ \hline
200	& 2.85	& 2.98	& 0.595	& 0.593 \\ \hline
\end{tabular}\label{tab:table1}
\end{table}

\subsection{Performance of Writer Identification} 
\label{perf_WI}

In this and next subsections, we present the {WI} and {WV} performances of the proposed system. 
Our system was evaluated with the same experimental strategy of \cite{10}.

For {WI/WV}, a 9-tuple accuracy measure obtained from various experimental setups was used in \cite{10}. 
However, in this 9-tuple, 3 elements computed the actual system performance for intra-variable handwriting inspection, i.e., training/testing on highly disparate 
{writing styles}, 
which was termed as \emph{3-tuple accuracy}. Therefore, in this paper, we focused on this 3-tuple accuracy measure to evaluate {WI/WV} performance. The 3-tuple accuracy was ($AE_{smv}, AE_{sfv}, AE_{mfv}$). $AE_{smv}$ was the average accuracy obtained from experimental setups $E_{sm}$ and $E_{ms}$. In $E_{sm}$ setup, the training was performed on $S_{s1}$ and testing was done on $S_{m3}$, i.e., $S_{s1}/S_{m3}$, while employing $D_c$. The $E_{ms}$ was the reverse experimental setup, i.e., $S_{m1}/S_{s3}$, when using $D_c$. As a matter of fact, on $D_{uc}$, $E_{sm}$ was $S_{s1}’/S_{m3}’$, and $E_{ms}$ was $S_{m1}’/S_{s3}’$. Similarly, $AE_{sfv}$ and $AE_{mfv}$ were obtained. 
A more detailed discussion on the experimental setup can be found in \cite{10}.


During the training of our system, 
{
some hyper-parameters were tuned and fixed.
We fixed 
momentum = $0.9$, 
learning rate = $0.01$, 
weight decay = $10^{-4}$
and 
epoch = $1000$.} 

For {WI}, the standard Top-N criterion was chosen, where we computed Top-1, Top-2, and Top-5 accuracy measure \cite{10}. As mentioned earlier, the {WI} task can be seen as a multi-class classification problem and we present the results in terms of accuracy.

In TABLE \ref{tab:table2}, we present the Top-1 {WI} performance in terms of 3-tuple accuracy using Inception-ResNet-v2 \cite{12} as I-net. 
From TABLE \ref{tab:table1}, we noticed the lowest MAE and the highest MIS were obtained for $k=125$ and $k=100$, respectively. 
Therefore, here 
we varied the $k$ in a smaller span from $75$ to $150$ with a step of $25$. Overall, we obtained the best performance for $k=100$ on both databases $D_c$ and $D_{uc}$. 
For $k=125$, the overall performance was the second-best, which was very close to the best. In general, the best 3-tuple accuracies for $D_c$ and $D_{uc}$ databases were (88.37\%, 81.57\%, 84.51\%) and (87.25\%, 79.67\%, 82.61\%), respectively. On $D_{uc}$, the accuracy $AE_{sfv}$ was slightly better for $k=125$ than for $k=100$. Overall, the intra-variable {WI} performance on $D_c$ was better than $D_{uc}$.    

\begin{table}
\scriptsize 
\caption{\small Top-1 writer identification performance}
\centering
\begin{tabular}{c|c|c|c|c}
\hline
\multirow{2}{*}{\textbf{DB}}	& \textbf{\# patches}	& \multicolumn{3}{c}{\textbf{3-tuple accuracy (\%)}} \\ \cline{3-5}
 & ($k$) &		$AE_{smv}$ &	$AE_{sfv}$ &	$AE_{mfv}$ \\ \hline \hline
\multirow{4}{*}{$D_c$} &	75	& 81.72	& 74.83	& 77.92 \\ \cline{2-5}
&	\textbf{100}	& \textbf{88.37} &	\textbf{81.57} &	\textbf{84.51} \\ \cline{2-5}
&	125 &	87.75 &	81.21 &	83.83 \\ \cline{2-5}
&	150 &	84.58 &	77.16 &	80.53 \\ \hlineB{2.5}
\multirow{4}{*}{$D_{uc}$} &	75 &	81.95 &	74.49 &	78.13 \\ \cline{2-5}
&	\textbf{100}	& \textbf{87.25} & 	79.67 & 	\textbf{82.61} \\ \cline{2-5}
&	125	& 86.74	& \textbf{79.69}	& 82.53 \\ \cline{2-5}
&	150	& 82.86	& 76.15	& 78.64 \\ \hline
\end{tabular}\label{tab:table2}
\end{table}

In general, our {WI} system performed the best when $100$ patches were selected from a text sample. Therefore, in this paper, we fix $k=100$, for presenting the rest of the experiments on intra-variable {WI}.

In TABLE \ref{tab:table3}, we present Top-1, Top-2, and Top-5 {WI} performances for $k=100$ with Inception-ResNet-v2 as I-net. 
The Top-2 measure was very close to the Top-1 performance. On $D_c$ and $D_{uc}$ databases, the Top-5 3-tuple accuracies were (91.54\%, 84.28\%, 87.18\%) and (91.08\%, 83.65\%, 87.10\%), respectively.

\begin{table}
\scriptsize 
\caption{\small Top-N writer identification with $k=100$}
\centering
\begin{tabular}{c|c|c|c|c}
\hline
\multirow{2}{*}{\textbf{DB}}	& \multirow{2}{*}{\textbf{Top-N}}	& \multicolumn{3}{c}{\textbf{3-tuple accuracy (\%)}} \\ \cline{3-5}
 &  &		$AE_{smv}$ &	$AE_{sfv}$ &	$AE_{mfv}$ \\ \hline \hline
\multirow{3}{*}{$D_c$}	& Top-1	& 88.37	& 81.57	& 84.51 \\ \cline{2-5}
	& Top-2	& 88.63	& 81.84	& 85.01 \\ \cline{2-5}
	& Top-5	& 91.54	& 84.28	& 87.18 \\ \hlineB{2.5}
\multirow{3}{*}{$D_{uc}$}	& Top-1	& 87.25 &	79.67 &	82.61 \\ \cline{2-5}
	& Top-2	& 88.13	& 80.81	& 83.33 \\ \cline{2-5}
	& Top-5	& 91.08	& 83.65	& 87.10 \\ \hline
\end{tabular}\label{tab:table3}
\end{table}

Our {WI} architecture (refer to Fig. \ref{fig:fig4}) is quite generalized, where we can employ various deep-feature generators as I-net. Apart from using the front part of the Inception-ResNet-v2 as I-net, we checked with the front part (up to "average pooling" layer) of some other powerful deep architectures, e.g., Inception-v4 \cite{12}, Xception net \cite{11}, as I-net. 
In TABLE \ref{tab:table4}, we present the Top-1 {WI} performance with various I-nets when $k=100$. 
Overall, we attained the best 3-tuple accuracies of (88.37\%, 81.57\%, 84.51\%) and (87.25\%, 79.67\%, 82.61\%) on $D_c$ and $D_{uc}$ databases, respectively, by employing Inception-ResNet-v2. The Inception-v4 performed similarly well, and became overall the second-best. The Xception net also performed quite well, though secured the last rank. The Xception net performed better than some state-of-the-art deep neural nets, e.g., Inception v3, Inception v2, GoogLeNet (Inception v1), VGG-16, ResNet-101, SqueezeNet, etc. \cite{10, 11, 2}.

\begin{table}
\scriptsize 
\caption{\small Top-1 writer identification on $k=100$ with various I-nets}
\centering
\begin{tabular}{c|c|c|c|c}
\hline
\multirow{2}{*}{\textbf{DB}}	& \multirow{2}{*}{\textbf{I-net}}	& \multicolumn{3}{c}{\textbf{3-tuple accuracy (\%)}} \\ \cline{3-5}
 &  &		$AE_{smv}$ &	$AE_{sfv}$ &	$AE_{mfv}$ \\ \hline \hline
\multirow{3}{*}{$D_c$} &	Inception-ResNet-v2 \cite{12} &	\textbf{88.37} &	\textbf{81.57}	& 84.51  \\ \cline{2-5}
	& Inception-v4 \cite{12}	& 88.20	& 81.53	& \textbf{84.52}  \\ \cline{2-5}
	& Xception net \cite{11}	& 87.76	& 80.72	& 83.96 \\ \hlineB{2.5}
\multirow{3}{*}{$D_{uc}$} &	Inception-ResNet-v2 \cite{12} &	\textbf{87.25} &	79.67 & 	\textbf{82.61} \\ \cline{2-5}
	& Inception-v4 \cite{12}	& 87.21	& \textbf{79.71}	& 82.53 \\ \cline{2-5}
	& Xception net \cite{11}	& 86.41	& 79.06	& 82.00 \\ \hline
\end{tabular}\label{tab:table4}
\end{table}

\subsection{Performance of Writer Verification}
\label{perf_WV}

Here also, we used the 3-tuple accuracy measure obtained from a similar experimental setup for {WI} of the previous subsection. The tuning of hyper-parameters was also similar to the {WI}. A small difference lies in measuring the accuracy, which has been discussed in Section \ref{4WV}. 

As mentioned earlier, {WV} can be perceived as a binary classification task to decide two handwriting samples either as "same" or "different" based on a given text sample. 
We present the results here in terms of accuracy (balanced), as given in Section \ref{4WV}. 

For {WV}, the features obtained from the patches of a text sample was aggregated using two different algorithms, i.e., MAF ({Mean Aggregated Feature}) and XAF ({maX Aggregated Feature}). We compare these two algorithms in TABLE \ref{tab:table5}, where we present the 3-tuple accuracy for {WV} on a variable number of patches ($k$) on databases $D_c$ and $D_{uc}$. Inception-ResNet-v2 was used here as I-net, and the triplet network was used for similarity learning.

\begin{table}[!b]
\scriptsize 
\caption{\small Writer verification performance}
\centering
\begin{tabular}{c|c|c|c|c|c}
\hline
\multirow{2}{*}{\textbf{DB}} & \multirow{2}{*}{\textbf{Algo}}	& \textbf{\# patches}	& \multicolumn{3}{c}{\textbf{3-tuple accuracy (\%)}} \\ \cline{4-6}
 &  & ($k$) &		$AE_{smv}$ &	$AE_{sfv}$ &	$AE_{mfv}$ \\ \hline \hline
\multirow{8}{*}{$D_c$}	& \multirow{4}{*}{MAF}	& 75 &	88.24 &	80.63 &	85.88 \\ \cline{3-6}
		& & \textbf{100}	& \textbf{94.87} &	\textbf{86.78} &	\textbf{92.12} \\ \cline{3-6}
		& & 125	& 94.62 &	86.32 &	91.52 \\ \cline{3-6}
		& & 150	& 92.83 &	84.18 &	89.01 \\ \cline{2-6}
	& \multirow{4}{*}{XAF}	& 75	& 83.76 &	75.59 &	81.72 \\ \cline{3-6}
		& & \textbf{100}	& \textbf{90.71} &	\textbf{82.47} &	\textbf{87.93} \\ \cline{3-6}
		& & 125	& 90.49 &	82.21 &	87.18 \\ \cline{3-6}
		& & 150	& 87.64 &	80.26 &	85.15 \\ \hlineB{2.5}
\multirow{8}{*}{$D_{uc}$}	& \multirow{4}{*}{MAF}	& 75 &	84.59 &	79.00 &	83.09 \\ \cline{3-6}
		& & \textbf{100}	& \textbf{92.45} &	86.35 &	\textbf{90.97} \\ \cline{3-6}
		& & 125	& 91.76 &	\textbf{86.43} &	90.63 \\ \cline{3-6}
		& & 150	& 90.63 &	85.25 &	89.20 \\ \cline{2-6}
	& \multirow{4}{*}{XAF}	& 75	& 81.78 &	75.40 &	80.19 \\ \cline{3-6}
		& & \textbf{100}	& \textbf{88.23} &	\textbf{81.86} &	\textbf{86.46} \\ \cline{3-6}
		& & 125	& 88.05 &	81.34 &	86.17 \\ \cline{3-6}
		& & 150	& 85.36 &	79.33 &	83.91 \\ \hline
\end{tabular}\label{tab:table5}
\end{table}

From TABLE \ref{tab:table5}, we note that overall we obtained the best performance for $k=100$ using both MAF and XAF algorithms on $D_c$ and $D_{uc}$. Therefore, in this paper, we used $k=100$ for presenting the rest of the experiments for intra-variable {WV}. Comparing MAF and XAF, we observed that MAF worked better than XAF on a variable number of patches for both the databases $D_c$ and $D_{uc}$. On $D_c$ and $D_{uc}$, the overall best 3-tuple accuracies were (94.87\%, 86.78\%, 92.12\%) and (92.45\%, 86.35\%, 90.97\%) using MAF, while $k=100$. In general, the intra-variable {WV} performance on $D_c$ was better than $D_{uc}$.

Similar to TABLE \ref{tab:table4} for {WI}, here in TABLE \ref{tab:table6}, we present the {WV} performance with various I-nets by employing MAF, triplet net and $k=100$. 
We obtained here the best 3-tuple accuracy (94.87\%, 86.78\%, 92.12\%) and (92.45\%, 86.35\%, 90.97\%) using Inception-ResNet-v2 on $D_c$ and $D_{uc}$, respectively. The results, employing Inception-v4 were very close to the best performance.

\begin{table}
\scriptsize 
\caption{\small Writer verification with various I-nets}
\centering
\begin{tabular}{c|c|c|c|c}
\hline
\multirow{2}{*}{\textbf{DB}}	& \multirow{2}{*}{\textbf{I-net}}	& \multicolumn{3}{c}{\textbf{3-tuple accuracy (\%)}} \\ \cline{3-5}
 &  &		$AE_{smv}$ &	$AE_{sfv}$ &	$AE_{mfv}$ \\ \hline \hline
\multirow{3}{*}{$D_c$}	& Inception-ResNet-v2 \cite{12}	& \textbf{94.87}	& \textbf{86.78}	& \textbf{92.12} \\ \cline{2-5}
	& Inception-v4 \cite{12}	& 94.71 &	86.34 &	91.83 \\ \cline{2-5}
	& Xception \cite{11}	& 93.66	& 85.19	& 90.05 \\ \hlineB{2.5}
\multirow{3}{*}{$D_{uc}$}	& Inception-ResNet-v2 \cite{12}	& \textbf{92.45}	& \textbf{86.35}	& \textbf{90.97} \\ \cline{2-5}
	& Inception-v4 \cite{12}	& 91.86	& 86.02	& 90.63 \\ \cline{2-5}
	& Xception \cite{11}	& 90.39	& 84.06	& 89.17 \\ \hline
\end{tabular}\label{tab:table6}
\end{table}

In TABLE \ref{tab:table7}, we present the {WV} performance with various similarity learning using MAF, Inception-ResNet-v2, and $k=100$. 
The triplet network worked better than the Siamese net, and produced 3-tuple accuracies of (94.87\%, 86.78\%, 92.12\%) and (92.45\%, 86.35\%, 90.97\%) on $D_c$ and $D_{uc}$, respectively.

\begin{table}
\scriptsize 
\caption{\small Writer verification with various similarity learning}
\centering
\begin{tabular}{c|c|c|c|c}
\hline
\multirow{2}{*}{\textbf{DB}}	& \multirow{2}{*}{\textbf{Similarity learning}}	& \multicolumn{3}{c}{\textbf{3-tuple accuracy (\%)}} \\ \cline{3-5}
 &  &		$AE_{smv}$ &	$AE_{sfv}$ &	$AE_{mfv}$ \\ \hline \hline
\multirow{2}{*}{$D_c$}	& Triplet net \cite{16}	& \textbf{94.87}	& \textbf{86.78}	& \textbf{92.12} \\ \cline{2-5}
	& Siamese net \cite{18}	& 88.95	& 81.59	& 86.78 \\ \hlineB{2.5}
\multirow{2}{*}{$D_{uc}$} &	Triplet net \cite{16} &	\textbf{92.45} &	\textbf{86.35} &	\textbf{90.97} \\ \cline{2-5}
	& Siamese net \cite{18}	& 87.40	& 80.63	& 84.46 \\ \hline
\end{tabular}\label{tab:table7}
\end{table}

From the above experiments on writer inspection, our major 
{observations} 
are summarized as follows.

{{\bf{\em{(i)}}} The {WI/WV} performance on $D_c$ database was better than on $D_{uc}$.

{{\bf{\em{(ii)}}} For {WI/WV}, the overall best performance was obtained while $100$ ($= k$) {IdS} patches per text sample were used. 

{{\bf{\em{(iii)}}} For {WI/WV}, the front part of the Inception-ResNet-v2 (up to "average pooling" layer) worked the best as I-net. 

{{\bf{\em{(iv)}}} Overall, the {WI/WV} performances in terms of individual elements of 3-tuple accuracy in decreasing order were as follows: $AE_{smv} \succ AE_{mfv} \succ AE_{sfv}$.

{{\bf{\em{(v)}}} For {WV}, in general, the MAF algorithm worked better than XAF. 

{{\bf{\em{(vi)}}} For {WV}, the triplet network worked better than the Siamese net for similarity learning.

\subsection{Comparison}
\label{compare}


{In this subsection, we compare our method with some major related work reported in the literature.}
We first compare with respect to idiosyncrasy analysis, then {WI} followed by {WV}.

\subsubsection{{Comparison of Idiosyncrasy Analysis}}
To compare our method of idiosyncrasy analysis, we came across only one work \cite{3} reported in the literature. 

In this paper, our task is to predict the idiosyncratic ({IdS}) score from a patch using deep regression analysis. Adak et al. \cite{3} modeled the task into classification problem to classify the text-patches into some highly {IdS} classes, i.e., $ID_1$ class when normalized score $i_j$ of patch $p_j$ was in the interval $(0.9, 1]$, $ID_2$ when $i_j$ was in the range $(0.8, 0.9]$, $ID_3$ when $i_j$ was in $(0.7, 0.8]$, and so on. For comparison purposes, we did a similar setting here, i.e., if $i_j$ lied in $(0.9, 1]$, then $p_j$ was in class $ID_1$, and so forth, to be in $ID_2$ and $ID_3$. Here, if the actual score $i_j$ of $p_j$ was in $ID_1$, and the regression-based predicted score ($\not=$ actual score) of $p_j$ was also in $ID_1$, then the $p_j$ was correctly classified (true positive), where the relaxed error was less than $0.1$. For comparison, we calculated the accuracy (balanced) \cite{21} from the top three {IdS} classes ($ID_1$, $ID_2$, and $ID_3$, i.e., $ID_{1-3}$), since $ID_{1-3}$ produced the best performance in \cite{3}. 
The quantitative comparison measure, using $100$ patches obtained from each text sample of intra-variable databases ($D_c$ and $D_{uc}$), is presented in TABLE \ref{tab:table8}.
%
{From this table}, we observe that our proposed method produced 98.35\% and 97.74\% accuracies on $D_c$ and $D_{uc}$ databases, respectively, which were better than the performances obtained in \cite{3}.

\begin{table}
\scriptsize 
\caption{\small Comparison of idiosyncrasy analysis}
\centering
\begin{tabular}{c|c|c}
\hline
\textbf{DB}	&  \textbf{Method}	& \textbf{Accuracy (\%)} \\ \hline \hline
\multirow{2}{*}{$D_c$}	& Adak et al. \cite{3}	& 90.56 \\ \cline{2-3}
	& Proposed	& \textbf{98.35} \\ \hlineB{2.5}
\multirow{2}{*}{$D_{uc}$} &	Adak et al. \cite{3}	& 89.85 \\ \cline{2-3}
	& Proposed	& \textbf{97.74} \\ \hline
\end{tabular}\label{tab:table8}
\end{table}

\subsubsection{Comparison of Writer Identification}
%
%
{
For comparative analysis concerning {WI} and  {WV} from intra-variable handwriting, we maintained similar experimental setups, employed databases, and performance measures as used in this paper (refer to Section \ref{DB_emp}), which is the same as in \cite{10}.}

{
An empirical study on intra-variable writer inspection was presented in \cite{10}, where the XceptionNet-based method "XN\_allo\_mean" performed better than some major state-of-the-art deep feature-based architectures (e.g., SqueezeNet, GoogLeNet, VGG-16,  ResNet-101, etc.) and some handcrafted feature-based models.
Therefore, from \cite{10}, we compared only with the XN\_allo\_mean method.
}

Another work reported in \cite{3} was on analyzing {IdS} handwriting to identify a writer. 
However, they did not work on intra-variable writing. Therefore, here, we were interested in executing the method of \cite{3} on our intra-variable writing's experimental setup. 

{
The state-of-the-art writer inspection methods did not focus on intra-variable handwriting. 
However, we intended to see the performance of the past methods on intra-variability. 
Therefore, we performed a comparative study with the past significant handcrafted \cite{2002Srihari, 23, COLD, SIFT} and deep feature-based \cite{CNN_Fiel, CNN_Christlein, CNN_Tang_Wu, He_deep,  3, 10} methods.
}

\begin{table}
\scriptsize 
\caption{\small {Comparison of Top-1 writer identification}}
\centering
\begin{tabular}{c|l|c|c|c}
\hline
\multirow{2}{*}{\textbf{DB}}	& \multirow{2}{*}{\textbf{Method}}	& \multicolumn{3}{c}{\textbf{3-tuple accuracy (\%)}} \\ \cline{3-5}
 &  &		$AE_{smv}$ &	$AE_{sfv}$ &	$AE_{mfv}$ \\ \hline \hline
\multirow{11}{*}{$D_c$}	
& {Macro-micro} \cite{2002Srihari} & 37.81 &	30.47 &	35.63 \\ \cline{2-5}
& {Hinge} \cite{23} 	& 46.56 & 39.95 & 44.16 \\ \cline{2-5}
& {LBPruns\_G + COLD} \cite{COLD}	& 49.13 & 42.66 & 46.47 \\ \cline{2-5}
& {SIFT (SDS + SOH)} \cite{SIFT}  & 52.36 & 45.61 & 48.55 \\ \cline{2-5}
& {CaffeNet} \cite{CNN_Fiel}  & 65.74 & 55.36 & 61.11 \\ \cline{2-5}
& {CNN + GMM} \cite{CNN_Christlein}   & 66.57 & 56.17 & 61.49 \\ \cline{2-5}
& {CNN + Bayesian} \cite{CNN_Tang_Wu}  & 68.29 & 58.35 & 63.24 \\ \cline{2-5}
& {Adaptive CNN + WLE} \cite{He_deep}  & 69.72 & 59.53 & 65.06 \\ \cline{2-5}
& XceptionNet (XN\_allo\_mean) \cite{10} & 73.74 & 64.12 & 68.94 \\ \cline{2-5}
& Idiosyncrasy + SqueezeNet \cite{3}	& 75.42	& 66.75	& 71.04 \\ \cline{2-5}
& Proposed	& \textbf{88.37}	& \textbf{81.57}	& \textbf{84.51} \\ \hlineB{2.5}
\multirow{11}{*}{$D_{uc}$}	
& {Macro-micro} \cite{2002Srihari} 	& 37.01	& 29.77 & 33.92 \\ \cline{2-5}
& {Hinge} \cite{23} 	& 46.15 & 39.56 & 43.63 \\ \cline{2-5}
& {LBPruns\_G + COLD} \cite{COLD}	& 49.03 & 42.39 & 46.35 \\ \cline{2-5}
& {SIFT (SDS + SOH)} \cite{SIFT}  & 51.97 & 45.05 & 48.63 \\ \cline{2-5}
& {CaffeNet} \cite{CNN_Fiel}  & 64.52 & 53.89 & 59.54 \\ \cline{2-5}
& {CNN + GMM} \cite{CNN_Christlein}   & 65.23 & 54.93 & 60.82 \\ \cline{2-5}
& {CNN + Bayesian} \cite{CNN_Tang_Wu}  & 67.42 & 57.31 & 62.78 \\ \cline{2-5}
& {Adaptive CNN + WLE} \cite{He_deep}  & 68.86 & 58.92 & 64.53 \\ \cline{2-5}
& XceptionNet (XN\_allo\_mean) \cite{10}& 72.52	& 62.79	& 66.53 \\ \cline{2-5}
& Idiosyncrasy + SqueezeNet \cite{3} & 73.68 & 64.07 & 68.84 \\ \cline{2-5}
& Proposed	& \textbf{87.25} & \textbf{79.67} & \textbf{82.61} \\ \hline
\end{tabular}\label{tab:table9}
\end{table}


In TABLE \ref{tab:table9}, we compare our proposed {WI} model (employing $k=100$, and Inception-ResNet-v2 as I-net) with
{some crucial related work 
in terms of Top-1 3-tuple accuracy on databases $D_c$ and $D_{uc}$.
From this table, we can see that handcrafted feature-based models performed poorly. The SIFT-based architecture \cite{SIFT} performed the best among the handcrafted feature-based models. 
The deep feature-based methods obtained better results than the handcrafted feature-based models. 
He et al. \cite{He_deep} identified the writers based on a single word. 
For comparison, we executed their {WI} method on every segmented word, and decided the ensemble page-level writer by max-voting. WLE (Word Length Estimation) was chosen here as an auxiliary task.   
Our proposed model} 
performed the best, which attained (88.37\%, 81.57\%, 84.51\%) and (87.25\%, 79.67\%, 82.61\%) 3-tuple accuracies on $D_c$ and $D_{uc}$, respectively. 
{
The method of \cite{3} ranked the second-best by leveraging idiosyncrasy analysis and SqueezeNet.
The other past methods} 
did not focus on the idiosyncrasy of writing. This attests to the importance of {IdS} handwriting analysis for {intra-variable} writer inspection.


\subsubsection{Comparison of Writer Verification}
{
This subsection compares our proposed {WV} method with some major state-of-the-art handcrafted \cite{2002Srihari, 23, COLD, SIFT} and deep feature-based \cite{CNN_Fiel, CNN_Christlein, CNN_Tang_Wu, He_deep,  3, 10} techniques, similar to the {WI}.
The experimental setups, employed databases, and performance measures were the same as in \cite{10}.}

{
Some of the past methods \cite{COLD, SIFT, CNN_Fiel, CNN_Christlein, CNN_Tang_Wu, He_deep, 3} did not report working with {WV}. 
However, these methods can easily verify a writer by using some distance measure between the features extracted from the known and questioned handwriting samples \cite{23, 10}. 
Therefore, for a comparative study, we executed these past methods to verify the writer while employing the Euclidean distance measure.}

{
In \cite{10}, for {WV} on intra-variable writing, the XceptionNet-based XN\_allo\_mean performed better than some past handcrafted and deep feature-based methods.
Therefore, from \cite{10}, we compared our method with the XN\_allo\_mean only.}

{
In TABLE \ref{tab:table10}, we present the {WV} performance of some important past methods executed on databases $D_c$ and $D_{uc}$.
Here also, the deep features worked better than the handcrafted features. 
Our proposed method obtained the best result, when we used $k=100$, Inception-ResNet-v2 as I-net, MAF for feature aggregation, and the triplet network for similarity learning. 
We achieved (94.87\%, 86.78\%, 92.12\%) and (92.45\%, 86.35\%, 90.97\%) 3-tuple accuracies on the $D_c$ and $D_{uc}$ databases, respectively. The method of \cite{3} also attained the second-best intra-variable {WV} performance.}



 

\begin{table}
\scriptsize 
\caption{\small {Comparison of writer verification}}
\centering
\begin{tabular}{c|l|c|c|c}
\hline
\multirow{2}{*}{\textbf{DB}}	
& \multirow{2}{*}{\textbf{Method}}	& \multicolumn{3}{c}{\textbf{3-tuple accuracy (\%)}} \\ \cline{3-5}
&  &	$AE_{smv}$ &	$AE_{sfv}$ &	$AE_{mfv}$ \\ \hline \hline
\multirow{11}{*}{$D_c$} 
& {Macro-micro} \cite{2002Srihari} & 48.50 & 36.58 & 41.43 \\ \cline{2-5}
& {Hinge} \cite{23} 	& 55.97 & 47.15 & 51.98 \\ \cline{2-5}
& {LBPruns\_G + COLD} \cite{COLD}	& 59.62	 & 50.86 & 55.34 \\ \cline{2-5}
& {SIFT (SDS + SOH)} \cite{SIFT}  & 61.91 & 53.35 & 58.02 \\ \cline{2-5}
& {CaffeNet} \cite{CNN_Fiel}  & 75.74 & 64.75 & 68.95 \\ \cline{2-5}
& {CNN + GMM} \cite{CNN_Christlein}  	& 77.23 & 66.21 & 70.85 \\ \cline{2-5}
& {CNN + Bayesian} \cite{CNN_Tang_Wu}  & 78.07 & 67.18 & 71.46 \\ \cline{2-5}
& {Adaptive CNN + WLE} \cite{He_deep}  & 79.12 & 67.34 & 72.09 \\ \cline{2-5}
&	XceptionNet (XN\_allo\_mean) \cite{10}	& 80.79	& 70.02	& 74.98 \\ \cline{2-5}
& {Idiosyncrasy + SqueezeNet} \cite{3} & 83.45 & 72.97 & 78.74 \\ \cline{2-5}
& Proposed	& \textbf{94.87}	& \textbf{86.78}	& \textbf{92.12} \\ \hlineB{2.5}
\multirow{11}{*}{$D_{uc}$}	
& {Macro-micro} \cite{2002Srihari} & 47.96 & 35.76 & 40.26 \\ \cline{2-5}
& {Hinge} \cite{23} 	& 55.25 & 46.91 & 51.27 \\ \cline{2-5}
& {LBPruns\_G + COLD} \cite{COLD}	& 58.85 & 49.26 & 54.63 \\ \cline{2-5}
& {SIFT (SDS + SOH)} \cite{SIFT}  & 61.33 & 52.82 & 57.30 \\ \cline{2-5}
& {CaffeNet} \cite{CNN_Fiel}  & 75.36 & 64.07 & 68.75 \\ \cline{2-5}
& {CNN + GMM} \cite{CNN_Christlein}   & 76.77 & 65.42 & 70.63 \\ \cline{2-5}
& {CNN + Bayesian} \cite{CNN_Tang_Wu}  & 77.89 & 67.00 & 71.08 \\ \cline{2-5}
& {Adaptive CNN + WLE} \cite{He_deep}  & 78.66 & 67.12 & 71.94 \\ \cline{2-5}
& XceptionNet (XN\_allo\_mean) \cite{10} & 79.84	& 69.80 & 74.76 \\ \cline{2-5}
& {Idiosyncrasy + SqueezeNet} \cite{3} & 81.82 & 71.35 & 77.26 \\ \cline{2-5}
& Proposed	& \textbf{92.45}	& \textbf{86.35}	& \textbf{90.97} \\ \hline
\end{tabular}\label{tab:table10}
\end{table}

From this comparative study, we observed that our present method outperformed the past methods for writer inspection on intra-variable data. We also observed that idiosyncrasy analysis aided the {WI/WV} system to perform better on intra-variable handwriting.

\subsection{{Ablation Study}}
\label{ablation}
{
In this subsection, we validate the requirement of idiosyncratic (IdS) patch detection for writer inspection. 
Furthermore, we check the need for RL to find IdS patches. 
For this, we use the strategy of ablation study \cite{Ablation}, where we remove a module to see whether the removal has affected the overall system.}

\subsubsection{{Ablating Idiosyncratic Patch Detection}}

{
For writer inspection, our proposed deep feature-based method, with the top 100 IdS patches obtained using RL, performed better than other deep feature-based methods \cite{CNN_Fiel, CNN_Christlein, CNN_Tang_Wu, He_deep, 10} that did not employ the idiosyncrasy analysis (refer to Section \ref{compare}).} 

{
To check the requirement of highly IdS patches, we execute the following two methods that ablate the RL-based IdS patch detection module from our proposed writer inspection system: 
{\bf{\em(i)}} \emph{Method-Rand\_100} that uses 100 random patches obtained from a handwritten sample,  
{\bf{\em(ii)}} \emph{Method-All} that employs all 
patches acquired from a sample. 
These two methods are compared with our proposed {WI}/{WV} method that used 100 highly IdS patches.} 

%
%

\begin{table}[!b]
\scriptsize 
\caption{\small {Ablation study: idiosyncratic patch detection}}
\centering
\begin{tabular}{c|c|l|c|c|c}
\hline
\textbf{Writer} & \multirow{2}{*}{\textbf{DB}} & \multirow{2}{*}{\textbf{Method}} & \multicolumn{3}{c}{\textbf{3-tuple accuracy (\%)}} \\ \cline{4-6}
\textbf{Inspection} & &  &	$AE_{smv}$ &	$AE_{sfv}$ &	$AE_{mfv}$ \\ \hline \hline
\multirow{6}{*}{WI} & \multirow{3}{*}{$D_c$} & {Rand-100} & 63.53 & 52.54 & 58.33 \\ \cline{3-6}
& & {All} & 73.97 & 64.23 & 69.19 \\ \cline{3-6}
& & Proposed & \textbf{88.37} & \textbf{81.57} & \textbf{84.51} \\ \cline{2-6}
& \multirow{3}{*}{$D_{uc}$} & {Rand-100} & 61.99 & 50.93 & 56.80 \\ \cline{3-6}
& & {All} & 72.95 & 63.05 & 66.78 \\ \cline{3-6}
& & Proposed & \textbf{87.25} & \textbf{79.67} & \textbf{82.61} \\ \hlineB{2.5}
\multirow{6}{*}{WV} & \multirow{3}{*}{$D_c$} & {Rand-100} & 73.60 & 62.02 & 66.01 \\ \cline{3-6}
& & {All} & 80.73 & 70.32 & 75.30 \\ \cline{3-6}
&& Proposed	& \textbf{94.87} & \textbf{86.78} & \textbf{92.12} \\ \cline{2-6}
& \multirow{3}{*}{$D_{uc}$} & {Rand-100} & 72.52 & 61.64 & 66.63 \\ \cline{3-6}
& & {All} & 79.83 & 69.93 & 74.79 \\ \cline{3-6}
& & Proposed & \textbf{92.45} & \textbf{86.35} & \textbf{90.97} \\ \hline
\end{tabular}\label{tab:table11}
\end{table}

{
In TABLE \ref{tab:table11}, we present the comparative study where we used similar experimental setups as discussed in Sections \ref{perf_WI}, \ref{perf_WV}. 
All the {WI} and {WV} methods of TABLE \ref{tab:table11} used Inception-ResNet-v2 as I-net.  
All the {WV} methods of this table used MAF for feature aggregation, and a triplet network for similarity learning.  
We repeated the experiment of \emph{Method-Rand\_100} 30 times, and present the average result in this table. 
From TABLE \ref{tab:table11}, we can observe that our proposed method (with top 100 IdS patches) outperformed the \emph{Method-Rand\_100} (using 100 random patches) and \emph{Method-All} (using all patches of a sample) for the {WI} and {WV} on databases $D_c$ and $D_{uc}$. 
This validates the necessity of highly idiosyncratic patches in our task.}



\subsubsection{{Ablating Reinforcement Learning}}

Now, we check the requirement of the RL for highly IdS patch detection. 
In \cite{3}, Inception module \cite{inception}-based deep features were used for IdS patch detection which did not work well when compared with our RL-based model in TABLE \ref{tab:table8}.

{
Our RL strategy explores a handwritten sample to find the IdS patch location automatically by exploiting the gathered knowledge.  
To validate the requirement of RL, we ablate the RL-based automated patch-location finding strategy.  
Alternatively, we use a sliding window-based protocol which finds all the patches. We then employ ResNet-50 (up to the "{avg pool}" layer) \cite{4} to extract deep features from every patch, and perform a regression analysis to produce the corresponding IdS score. The top $k$ IdS score-based patches are chosen to inspect the writer. We refer to the method that ablates RL as \emph{Method-Ablate\_RL}.
In TABLE \ref{tab:table12}, we compare this method with our proposed method on $D_c$ and $D_{uc}$ using similar experimental setups as in Sections \ref{perf_WI}, \ref{perf_WV} when $k=100$. 
\emph{Method-Ablate\_RL} employed Inception-ResNet-v2 as I-net similar to our proposed method for {WI} and  {WV}. 
\emph{Method-Ablate\_RL} also used MAF for feature aggregation and a triplet network for similarity learning like our proposed method for {WV}.
From TABLE \ref{tab:table12}, we note that our proposed RL-based method performs better than \emph{Method-Ablate\_RL} for WI/WV. 
We observe \emph{Method-Ablate\_RL} restricts the independent finding of patch location. 
Furthermore, \emph{Method-Ablate\_RL} is around six times computationally slower than our proposed RL-based method due to locating all patches in the initial stage, when tested on $D_c$ and $D_{uc}$. 
This signifies the requirement of RL for idiosyncrasy analysis.}

\begin{table}[!htb]
\scriptsize 
\caption{\small {Ablation study: reinforcement learning}}
\centering
\begin{tabular}{c|c|l|c|c|c}
\hline
\textbf{Writer} & \multirow{2}{*}{\textbf{DB}} & \multirow{2}{*}{\textbf{Method}} & \multicolumn{3}{c}{\textbf{3-tuple accuracy (\%)}} \\ \cline{4-6}
\textbf{Inspection} & &  &	$AE_{smv}$ &	$AE_{sfv}$ &	$AE_{mfv}$ \\ \hline \hline
\multirow{4}{*}{WI} & \multirow{2}{*}{$D_c$} & {Ablate\_RL} & 80.94 & 72.37 & 76.96  \\ \cline{3-6}
& & Proposed & \textbf{88.37} & \textbf{81.57} & \textbf{84.51} \\ \cline{2-6}
& \multirow{2}{*}{$D_{uc}$} & {Ablate\_RL} & 79.14 & 70.94 & 75.74 \\ \cline{3-6}
& & Proposed & \textbf{87.25} & \textbf{79.67} & \textbf{82.61} \\ \hlineB{2.5}
\multirow{4}{*}{WV} & \multirow{2}{*}{$D_c$} & {Ablate\_RL} & 88.01 & 77.52 & 83.12 \\ \cline{3-6}
&& Proposed	& \textbf{94.87} & \textbf{86.78} & \textbf{92.12} \\ \cline{2-6}
& \multirow{2}{*}{$D_{uc}$} & {Ablate\_RL} & 86.33 & 75.96 & 82.47 \\ \cline{3-6}
& & Proposed & \textbf{92.45} & \textbf{86.35} & \textbf{90.97} \\ \hline
\end{tabular}
\label{tab:table12}
\end{table}




%% file: 6conclusion.tex
\section{Conclusion}
\label{6conclusion}

In this paper, we worked on {WI} and {WV} from intra-variable handwriting. Inspecting the writer's scribbling on a whole page did not produce good performance. Therefore, we first planned to detect some highly idiosyncratic patches, then performed the inspection from these patches. For such patch detection, we used a recurrent reinforcement learning-based technique, where the idiosyncratic score was predicted by deep feature-based regression analysis. {WI} and  {WV} were performed by deep neural architectures. We employed two databases $D_c$ and $D_{uc}$ for the experimental study. Our idiosyncrasy analyzer fostered a promising performance for the writer inspection system. For {WI}, we obtained the best Top-1 3-tuple accuracy of (88.37\%, 81.57\%, 84.51\%) and (87.25\%, 79.67\%, 82.61\%) on the $D_c$ and $D_{uc}$ databases, respectively. For {WV}, our system attained the best 3-tuple accuracy of (94.87\%, 86.78\%, 92.12\%) and (92.45\%, 86.35\%, 90.97\%) on $D_c$ and $D_{uc}$, respectively.

In the future, we will endeavor to generate the intra-variable writing synthetically, so that our system can learn various types of possible intra-variability of individual handwriting. Moreover, we will try to explore some implicit characteristics of handwritten strokes which may not change drastically due to intra-variability.

%% file: appendix.tex
\begin{appendices}
\label{app}

\section{Algorithms}


Refer to Section V and Fig. 5.


\begin{algorithm}
\caption{MAF: \emph{page\_level\_Mean\_Aggregated\_Feature}}
\begin{algorithmic}[1]
\State	Input: $p_j': \{p_1', p_2', \dots, p_k'\}~|$ top-$k$ idiosyncratic patches in a page image $H$;
\State	Output: $v^{(\mu)}: \{v_1^\mu,v_2^\mu, \dots, v_q^\mu\}~|$ a $q$-dimensional feature vector representing page image $H$;
 \For	{$j = 1 ~to~ k$}
 \State   $ v^{(j)} = \mhyphen{I-net}(p_j')$;  
\Comment{\textcolor{gray}{Comment: $v^{(j)} := \{v_1^j, v_2^j, \dots, v_q^j \} := \{v_x^j; ~\forall x = 1, 2, \dots q\}$}}
	\EndFor
 \State	$v^{(\mu)}=NULL$;
	\For {$x = 1 ~to~ q$}
	       \State $Sum = 0$;
	     \For {$j=1 ~to~ k$}
              \State $Sum = Sum + v_x^j$; 
	     \EndFor
	       \State $v_x^\mu = Sum/k$;
	       \State $v^{(\mu)}=v^{(\mu)}.~append(v_x^\mu)$;
	\EndFor
\State return $v^{(\mu)}$; 
\end{algorithmic}
\label{algo:algo1}
\end{algorithm}


\begin{algorithm}
\caption{XAF: \emph{page\_level\_maX\_Aggregated\_Feature}}
\begin{algorithmic}[1]
\State	Input: $p_j': \{p_1', p_2', \dots, p_k'\}~|$ top-$k$ idiosyncratic patches in a page image $H$;
\State	Output: $v^{(\mu)}: \{v_1^\mu,v_2^\mu, \dots, v_q^\mu\}~|$ a $q$-dimensional feature vector representing page image $H$;
 \For	{$j = 1 ~to~ k$}
 \State   $ v^{(j)} = \mhyphen{I-net}(p_j')$; 
\Comment{\textcolor{gray}{Comment: $v^{(j)} := \{v_1^j, v_2^j, \dots, v_q^j \} := \{v_x^j; ~\forall x = 1, 2, \dots q\}$}}
 \EndFor
 \State	$v^{(\mu)}=NULL$;
	\For {$x = 1 ~to~ q$}
	       \State $ Max = v_x^1 $;
	     \For {$j = 2 ~to~ k$}
	              \If   {$v_x^j > Max$}
	                   \State $Max = v_x^j$; 
	              \EndIf
	     \EndFor
	       \State $v_x^\mu = Max$;
	       \State $v^{(\mu)} = v^{(\mu)}.~append(v_x^\mu)$;
	\EndFor
\State return $v^{(\mu)}$; 
\end{algorithmic}
\label{algo:algo2}
\end{algorithm}

\end{appendices}